\documentclass{article} 
\usepackage{iclr2026_conference,times}
\iclrfinalcopy
\usepackage{amsmath}      
\usepackage{amssymb}      
\usepackage{amsthm}       
\usepackage{mathtools}    
\usepackage{bm}           
\usepackage{graphicx}   
\newtheorem{theorem}{Theorem}
\newtheorem{proposition}{Proposition}

\theoremstyle{definition}

\newtheorem{assumption}{Assumption}
\newtheorem{remark}{Remark}
\usepackage{wrapfig}
\usepackage{graphicx}     
\usepackage{booktabs}     
\usepackage{multirow}     
\usepackage{subcaption}   
\usepackage{array}        
\usepackage{adjustbox}    
\usepackage[table]{xcolor}
\usepackage{pgfplots}
\pgfplotsset{compat=1.18}

\usepackage{algorithm}
\usepackage{algpseudocode}   

\usepackage{url}
\usepackage[hidelinks]{hyperref}  
\usepackage{cleveref}     

\usepackage{xcolor}       
\usepackage{enumitem}     
\usepackage{microtype}    

\usepackage{amsmath,amsfonts,bm}

\def\eqref#1{equation~\ref{#1}}

\def\1{\bm{1}}

\DeclareMathAlphabet{\mathsfit}{\encodingdefault}{\sfdefault}{m}{sl}
\SetMathAlphabet{\mathsfit}{bold}{\encodingdefault}{\sfdefault}{bx}{n}

\DeclareMathOperator*{\argmax}{arg\,max}

\usepackage{hyperref}
\usepackage{url}

\title{Covariance-Aware Scoring: A Geometric Correction for Vision-Language Test-Time Adaptation}

\author{Lakpa Tamang, Saman Forouzandeh, Kamal Berahmand, Wei Shao, Hamid Beigy,\\
\textbf{\& Mahdi Jalili}  \\
Department of Electrical and Electronic Engineering\\
RMIT University\\
Melbourne, Australia \\
\texttt{\{lakpa.tamang,saman.forouzandeh,kamal.berahmand,}\\
\texttt{wei.shao,hamid.beigy,mahdi.jalili\}@rmit.edu.au}
}

\begin{document}

\maketitle

\begin{abstract}
Test-time adaptation (TTA) enables vision--language models such as CLIP to adapt to
distribution shifts without labels or access to source data. Existing methods adapt through feature caching, prototype estimation, or the online accumulation of feature statistics, but most ultimately score a test feature against a class representative in the ambient embedding space, typically through cosine similarity. We identify this shared decision
rule as a limitation under anisotropic feature distributions, which means when a small number of directions account for a large fraction of feature variance, the score can be dominated
by high-variance directions that are not necessarily discriminative. In light of this, we propose
CARe, a covariance-aware scoring for TTA that replaces the ambient geometry with a
Mahalanobis geometry induced by the target feature second moment. Theoretically, we show that under a shared-covariance Gaussian model the
covariance-aware decision direction coincides with that of the Bayes-optimal linear
discriminant, with ambient-space scoring as the isotropic special case, and establish
that the potential benefit of covariance-aware scoring grows with feature anisotropy. Empirically, we show that CARe
improves adaptation accuracy over its base cosine scoring on both standard and fine-grained distribution-shift
benchmarks across two backbones. Consistent
with our analysis, the gains are largest on datasets whose features are most
anisotropic and vanish as the features approach isotropy.

\end{abstract}

\section{Introduction}
\label{sec:intro}

Vision--language models (VLMs) such as CLIP~\citep{radford2021learning} enable zero-shot image classification by matching image features with textual class representations. However, their performance can degrade under distribution shifts between pretraining and deployment. Test-time adaptation (TTA) addresses this challenge by adapting models online using only the unlabeled test stream, without access to source data or target labels~\citep{liang2025comprehensive,sheng2025r}. Recent TTA methods have explored a range of adaptation mechanisms, including entropy minimization \citep{wang2020tent}, confidence-based pseudo-labeling \citep{tan2025uncertainty}, feature caching \citep{niu2022efficient}, prototype estimation \citep{huang2026prototype}, and online accumulation of feature statistics \citep{guan2026statistics}. 
Despite these differences, most approaches ultimately evaluate a test feature against a class representative within the ambient embedding space, typically using cosine similarity.

\begin{figure}[ht!]
  \centering
    \vspace*{-4mm}\includegraphics[width=0.7\linewidth]{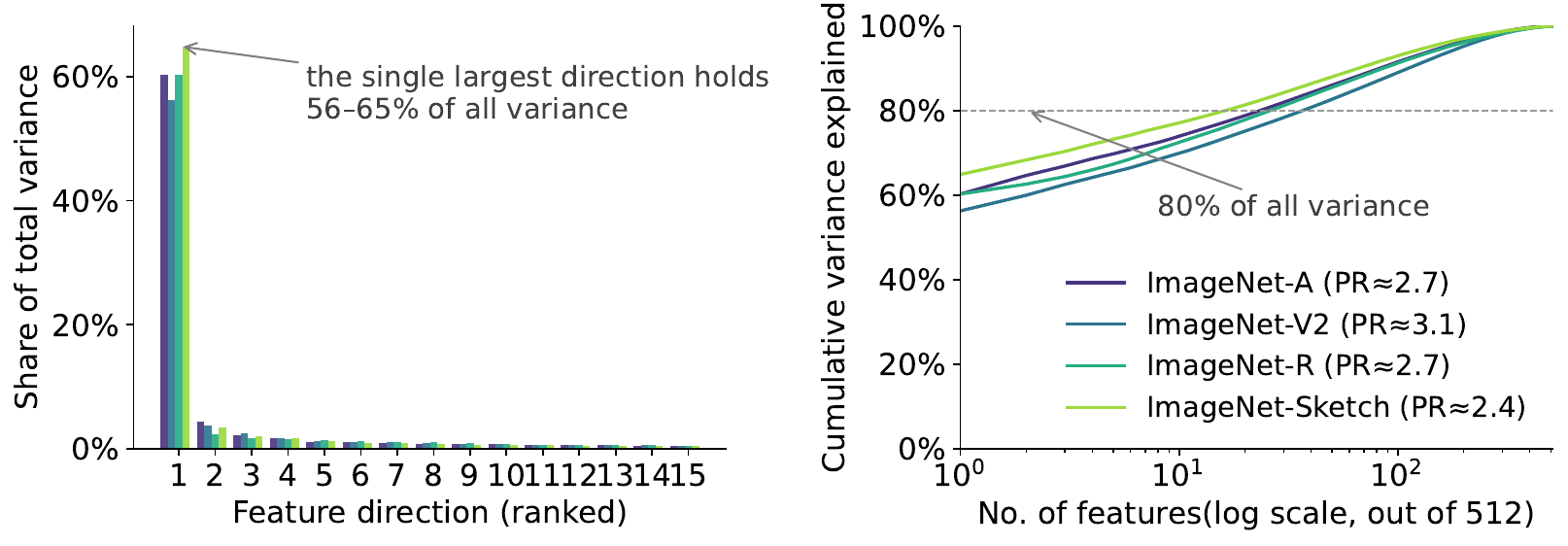}
    \vspace*{-3mm}
  \caption{\textbf{CLIP representations are highly anisotropic.} Across four Imagenet OOD benchmarks, the leading feature direction accounts for 56--65\% of the total variance, and approximately 80\% of the variance is concentrated in a small subset of the 512-dimensional space for ViT/B-16.}
  \label{fig:anisotropy}
\end{figure}

We argue that such a shared decision rule is an overlooked bottleneck in CLIP-based TTA. Ambient-space scoring implicitly treats all feature directions equally, whereas CLIP representations can exhibit strong anisotropy, with variance concentrated in a small number of directions. Figure~\ref{fig:anisotropy} illustrates this phenomenon across four out-of-distribution (OOD) benchmarks derived from ImageNet. Across ViT/B-16's 512-dimensional feature space, the leading direction alone explains
56\%--65\% of the total variance, and roughly 80\% of the variance is concentrated in fewer than 20 directions. The corresponding participation ratio (PR) is only 2.4--3.1, indicating an effective dimensionality of a handful of directions out of 512. Consequently, similarity scores can be disproportionately influenced by high-variance directions, even when class-discriminative information lies elsewhere. This observation exposes a limitation shared by existing TTA methods. Adaptation can improve the quality of features, class representations, or retrieved examples, but if the final prediction continues to compare them in the ambient geometry, the decision geometry itself remains unchanged.

We therefore propose \textbf{CARe} (\textbf{C}ovariance-\textbf{A}wa\textbf{Re} TTA), which adapts the decision geometry rather than the adaptation mechanism itself. One line of work which operate outside the ambient space is a probabilistic adapter
such as ADAPT~\citep{zhang2026backpropagation} that estimate class-conditional means and
a shared covariance and classify by Gaussian discriminant analysis. CARe differs from both approaches: unlike cosine-based adapters, it accounts for feature covariance; unlike probabilistic adapters, it uses only the uncentered total second moment accumulated from the test stream, without estimating class-conditional statistics or constructing a generative classifier. In particular, CARe accumulates second-order statistics from the test feature stream online and uses them to replace the Euclidean geometry underlying ambient-space scoring with a Mahalanobis geometry induced by the target second moment. We formulate CARe such that whenever the second moment of the features is isotropic, the scoring reduces exactly to the base cosine rule. We theoretically show this formulation connects empirical TTA to classical statistical decision theory. Under a shared-covariance Gaussian model, the covariance-aware decision direction coincides with that of the Bayes-optimal linear discriminant, while ambient-space scoring appears as the isotropic special case. The potential discrepancy between the two geometries increases with covariance anisotropy, providing a principled explanation of when CARe is most likely to help. Empirically, CARe improves its base adapter on both standard and fine-grained distribution-shift benchmarks with negligible computational overhead. These results suggest that improving TTA need not require a more elaborate adaptation procedure, and correcting the geometry used to compare adapted representations can itself yield consistent gains.
  \vspace*{-3mm}
\paragraph{Contributions.}
\begin{itemize}
    \item We identify a limitation common to most CLIP-based TTA methods: they retain a covariance-agnostic scoring rule that scores in the ambient space. This
    is consequential for CLIP, whose features are strongly anisotropic where the leading
    direction alone explains 56--65\% (Figure~\ref{fig:anisotropy}) of the variance.

    \item We propose CARe, a closed-form covariance-aware TTA method that scores under
    a Mahalanobis geometry induced by the second moment of test features. Our theoretical analysis characterizes
    when it helps and how its benefit scales with anisotropy (Section~\ref{sec:theory}).

    \item CARe improves over the existing baselines on standard and fine-grained benchmarks across two architectures, with gains tracking anisotropy on real data and in a controlled synthetic setting  (Section~\ref{sec:experiments}).
\end{itemize}

\section{Background and Preliminaries}
\label{sec:prelim}

\paragraph{Zero-shot CLIP.}
Let $\phi_I(\cdot)$ and $\phi_T(\cdot)$ denote the image and text encoders of a
pretrained CLIP model. For a $K$-class problem with classes $\{c_1,\dots,{c_K}\}$
and textual prompts $ {t}_k$, CLIP maps an image $ {x}\in\mathcal{X}$
and each prompt to $D$-dimensional embeddings,
$f=\phi_I( {x})\in\mathbb{R}^{D}$ and
$w_k=\phi_T( {t}_k)\in\mathbb{R}^{D}$. Writing
$\widehat{ {v}}= {v}/\lVert {v}\rVert_2$ for the $\ell_2$
normalization and the zero-shot class representatives (textual prompts) $ \bm{W}^{\text{Z.S}}=[ {w}^{\text{Z.S}}_1,\dots,{w^{\text{Z.S}}_K}]$ for the text
classifier, the zero-shot textual logit for class $k$ is the cosine similarity
$ {s}^{\text{Z.S}}_k=\cos({f}, {w}_k)$, and
the zero-shot posterior with temperature $\tau$ is
\begin{equation}
  p^{\text{Z.S}}_k = p(c_k\mid x)
  = \frac{\exp(z^{\text{Z.S}}_k)}{\sum_{j=1}^{K}\exp(z^{\text{Z.S}}_j)},
  \qquad z^{\text{Z.S}}_k = \frac{s^{\text{Z.S}}_k}{\tau}.
  \label{eq:zeroshot}
\end{equation}

\paragraph{Cache-based TTA.}
Cache-based methods~\citep{karmanov2024efficient} adapt the classifier without any gradient
updates by maintaining a small, class-balanced memory of at most $N$ entries per
class. As each test feature $f$ arrives, its zero-shot
prediction (Eq.~\ref{eq:zeroshot}) supplies a pseudo-label $\hat{y}$, and the pair
$(f,\hat y)$ is admitted to the cache under a minimal-entropy
criterion, such that the confident, and reliable exemplars are retained while the uncertain ones
discarded. Let $\mathcal{C}_k=\{f^{(k)}_1,\dots,f^{(k)}_{N_k}\}$
denote the cached features assigned to class $k$. At inference, the cache
contributes a similarity-based logit that is fused with the zero-shot logit:
\begin{equation}
   {z}_k
  \;=\;
  \alpha_1\cdot{z_k^\text{Z.S}}
  \;+\;
  \alpha_2\cdot\underbrace{\sum_{f^{(k)}_i\in\mathcal{C}_k}
      \sigma\!\big(f,f^{(k)}_i\big)}_{\text{cache similarity}},
  \label{eq:cache}
\end{equation}
where $\sigma(\cdot,\cdot)$ is a feature-similarity kernel (e.g.\ an exponential of
the cosine similarity) and $\alpha_1,\alpha_2$ balance the two terms.
 
\paragraph{Statistics-based adapters.}
This is a special case of cache-based TTA that caches sufficient statistics rather than features \citep{guan2026statistics}. From the stream it accumulates the second moment
$\bm{G}=\sum_i f_i f_i^{\top}\in\mathbb{R}^{D\times D}$ and the feature--label
cross-moment $\bm{P}=\sum_i q_i f_i^{\top}\in\mathbb{R}^{K\times D}$, where $q_i$ is a
soft pseudo-label derived from the zero-shot prediction, and obtains its class
representatives as the ridge solution
\begin{equation}
\bm{W}=(\bm{G}+\lambda_\text{ridge} \bm{I})^{-1}\bm{P}^{\top},
\label{eq:sca-w}
\end{equation}
where $\lambda_\text{ridge}>0$ is a ridge regularizer. Both moments ($\bm{G}$ and $\bm{P}$) have fixed size and absorb every observed sample, so the accumulated
knowledge does not decay with stream length. Note that $\bm{G}$ enters only through the construction of $\bm{W}$; the comparison
between $f$ and $\bm{W}$ still remains in the ambient geometry.

\section{Method}
\label{sec:method}

We consider the standard online TTA setting, where an unlabeled test stream $[ {x}_1, {x}_2,\dots]$ arrives sequentially. For each $ {x}_t$, a prediction must be made using only the samples observed up to that point, while the CLIP encoders remain fixed. From the observed test stream and a set of adapted class representatives $ \bm{W}\in\mathbb{R}^{D\times K}$ (from Eq.~\ref{eq:sca-w}) the prediction $\hat{y}({f})$ follows:
\begin{equation}
\hat{y}({f}) = \argmax_{k\in[K]} s_k^{\text{base}}({f, {w}_k}),
\qquad
s_k^{\text{base}}({f, {w}_k}) = \cos({f},\,{w}_k) = \frac{{f}^{\!\top}{w}_k}{\lVert{f}\rVert_2\,\lVert{w}_k\rVert_2}.
\label{eq:cosine-score}
\end{equation}

where $s_k^{\text{base}}$ is the cosine similarity between ${f}=\phi_I( {x})$ and $w_k$ (the $k$th column of $\bm{W}$).

\begin{figure}
    \centering
      \vspace*{-5mm}
      \includegraphics[width=\linewidth]{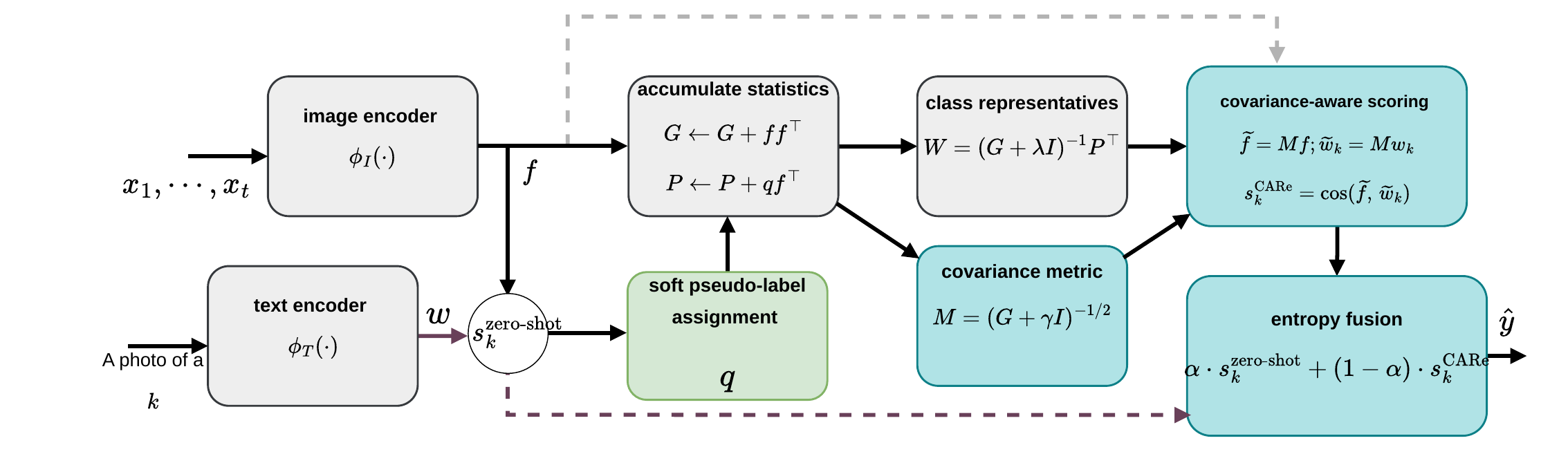}
    \vspace*{-5mm}
    \caption{Overview of CARe TTA. The encoded feature $f$ and a soft pseudo-label
$q$ derived from the zero-shot prediction update the statistics $\bm{G}$ and $\bm{P}$, from
which the class representatives $\bm{W}$ is created. These same
second moments also define the transformation $\bm{M}=(\bm{G}+\gamma I)^{-1/2}$, under which $f$ and
$\bm{W}$ are compared; the resulting score is fused with the zero-shot score.}
    \label{fig:care_overview}
\end{figure}

\subsection{Covariance-aware Scoring}
\label{sec:care}
CLIP representations are strongly anisotropic \citep{liang2022mind} (Figure~\ref{fig:anisotropy}), making ambient-space scoring potentially suboptimal: dominant high-variance directions can overwhelm lower-variance directions that are more discriminative (we formalize this effect in Section~\ref{sec:theory}). CARe addresses this by comparing $f$ and $w_k$ under a variance-aware metric that down-weights high-variance directions. The metric is estimated from second-order statistics of the test stream via a running $D\times D$ second moment, whose fixed size is independent of stream length and whose reliability improves as samples accumulate. In particular, we first accumulate the online running estimates of the uncentered second-moment matrix as follows:
\begin{equation}
\bm{G}_t \;=\; \sum_{i=1}^{t} f_i f_i^{\top},
\label{eq:secondmoment}
\end{equation}
where $t$ is the number of test samples seen so far. $\bm{G}_t$ is symmetric
positive-semidefinite and is updated online in $O(D^2)$ per step; we write
$\bm{G}$ for $\bm{G}_t$ when the time index is immaterial. It grows in scale with the length of the stream and is proportional to the second moment of the population $\mathbb{E}[ff^\top]$.
%
%
Using $\bm{G}$, we replace the ambient inner product with the Mahalanobis inner
product, which downweights high-variance directions according to their inverse
standard deviation and thereby balances their influence against directions that
may carry discriminative information. This is realized by the symmetric
positive-definite transform
\begin{equation}
  \bm{M} \;=\; \big(\bm{G}+\gamma\bm{I}\big)^{-1/2},
  \label{eq:transform}
\end{equation}
where $\gamma>0$ is a regularizer that governs how aggressively the second moment is inverted. In Eq.~\ref{eq:transform}, as
$\gamma\!\to\!\infty$, $\bm{M}$ tends to $\gamma^{-1/2}\bm{I}$ and
$s_k^{{CARe}}$ reduces exactly to the base scoring rule in
(Eq.~\ref{eq:cosine-score}), while $\gamma\!\to\!0^{+}$ {CARe} applies the
full Mahalanobis metric. We make this explicit theoretically through Remark \ref{rem:bayes-optimal} in Section~\ref{sec:theory}. Applying $\bm{M}$ to both the feature and the class
representatives before comparing them, {CARe} scores class $k$ via cosine
similarity in the transformed space,
\begin{equation}
  s_k^{\text{CARe}}(\widetilde{f}, \widetilde{{w}}_k)
  =
  \cos(\widetilde{{f}},\,\widetilde{{w}}_k)
  =
  \frac{(\bm{M} f)^\top (\bm{M} w_k)}{\lVert \bm{M }f\rVert_2 \lVert \bm{M} w_k \rVert_2}
  \label{eq:care-score}
\end{equation}
where $\widetilde{{f}}=\bm{M}{f}$ and
$\widetilde{{w}}_k=\bm{M}{w}_k$ denotes the transformed image feature
and class representative, respectively. The CARe posterior then follows:
\begin{equation}
  p^{\text{CARe}}_k
  = \frac{\exp(z^{\text{CARe}}_k)}{\sum_{j=1}^{K}\exp(z^{\text{CARe}}_j)},
  \qquad z^{\text{CARe}}_k = \frac{s^{\text{CARe}}_k}{\tau}
  \label{eq:careposterior}
\end{equation}
The schematic representation of adaptation using CARe is depicted in Figure~\ref{fig:care_overview}.

\paragraph{Pseudo-label assignment.}
Because no labels are available at test time, $\bm{P}$ is accumulated using
pseudo-labels derived from the zero-shot prediction. Rather than committing to the
single most confident class, we follow \citep{guan2026statistics} and retain a
small candidate set of plausible classes, sized adaptively by a
confidence-dependent threshold that admits more classes when the zero-shot
prediction is uncertain (the exact rule and its threshold $\tau_p$ are given in
Algorithm~\ref{alg:care}). The zero-shot posterior is restricted to this set and
renormalized to form the pseudo-label, so the target is sharp when the zero-shot
prediction is decisive and soft when uncertainty remains, reducing the impact of
erroneous hard assignments on the accumulated statistics.

\paragraph{Score fusion.}
We combine the covariance-aware score with the frozen zero-shot score using a sample-specific weight determined by their relative predictive entropies, rather than a single coefficient tuned per dataset.
\begin{equation}
\alpha
=
\frac{1}{1+\exp\!\left(
H^{\text{CARe}}-H^{\text{Z.S}}
\right)} \in\left[0, 1\right],
\qquad
\hat{y}
=
\arg\max_{k\in[K]}
\left[
(1-\alpha) s_k^{\text{Z.S}}
+
\alpha s_k^{\text{CARe}}
\right].
\label{eq:fusion}
\end{equation}
where $s_k^{\text{Z.S}}$ is the frozen zero-shot score, $H^{\text{CARe}}=\mathcal{H}(p_k^{\text{CARe}})$, and
$H^{\text{Z.S}}=\mathcal{H}(p_k^{\text{Z.S}})$;
$\mathcal{H}$ is Shannon entropy. This weighting is adaptive and computed directly from the predictive entropies, requiring no additional parameter tuning at test time.

\section{Theoretical Analysis}
\label{sec:theory}

Before establishing statistical optimality, we identify the exact geometric operation that  CARe  performs. Remark~\ref{rem:geometry} shows that the  CARe  score is not an \emph{ad hoc} similarity function but a standard cosine similarity measured in a data-dependent inner-product space.

\begin{remark}[CARe is cosine in a Mahalanobis geometry]
\label{rem:geometry}
Let $\bm{A}\succ{0}$ be any symmetric positive-definite matrix.
The Mahalanobis inner product and the induced norm are defined as
$\langle{u},{v}\rangle_{\bm{A}}:={u}^\top\bm{A}{v}$
and $\|{v}\|_{\bm{A}}:=\sqrt{\langle{v},{v}\rangle_{\bm{A}}}$
\citep{mahalanobis1936generalized}. Setting
$\bm{A}=(\bm{G}+\gamma \bm{I})^{-1}$,   the  CARe  score in
Eq.~\ref{eq:care-score} is exactly the cosine of the angle between $f$
and $w_k$ measured in this inner product:
\begin{equation}
s_k^{\text{CARe}}(f, w_k)
= \frac{\langle f,w_k\rangle_{\bm{A}}}{\|f\|_{\bm{A}}\|w_k\|_{\bm{A}}}
= \cos_{\bm{A}}(f,w_k).
\end{equation}
In the eigenbasis of $\bm{G}$, with eigenvalues $\lambda_1 \ge \lambda_2 \ge \cdots$,
the $i$-th eigendirection is down-weighted by $(\lambda_i+\gamma)^{-1}$. When $\bm{G}+\gamma \bm{I}\propto \bm{I}$,
$\cos_{\bm{A}}(f,w_k)$ reduces to the ordinary Euclidean angle and the score collapses
to the standard cosine rule Eq.~\ref{eq:cosine-score}.
\end{remark}

This identity reveals that  CARe  performs a per-eigendirection reweighting by $(\lambda_i+\gamma)^{-1}$, strongest along the high-variance directions that dominate ambient-space scoring. When $\bm{G}+\gamma \bm{I}\propto \bm{I}$, this reweighting is uniform and the correction is inactive.
To connect the uncentered scatter matrix $\bm{G}$ accumulated by  CARe  to the classical quantities of discriminant analysis, we posit the weakest generative model under which a linear scoring rule is Bayes-optimal.

\begin{assumption}[Shared-covariance Gaussian model]
\label{ass:shared-cov}
On the test distribution, class-conditional features are Gaussian with common
covariance, $f \mid y=k \sim \mathcal{N}(\mu_k, \bm{S})$ for all $k=1,\dots,K$, with
class priors $\pi_k$, and the class representatives align with the class means,
$w_k \propto \mu_k$.
\end{assumption}
The uncentered second moment decomposes as the shared covariance $\bm{S}$ plus a rank-$K$ mean term.
\begin{equation}
\bm{G} \;=\; \bm{S} \;+\; \sum_{k} \pi_k \mu_k \mu_k^{\top},
\label{eq:G-decomp}
\end{equation}


CARe accumulates $\bm{G}$, not $\bm{S}$, because $\bm{S}$ is unidentifiable without
labels. We use Assumption~\ref{ass:shared-cov} solely as an analytical lens: the
\emph{ideal} metric $\bm{S}^{-1}$ induces the Bayes decision direction
(Remark~\ref{rem:bayes-optimal}), while CARe uses $(\bm{G}+\gamma\bm{I})^{-1}$
because $\bm{S}$ is not accessible. The two are not equivalent: $\bm{G}$ also
contains the mean term (decomposed below), so inverting it is a trade-off rather
than a recovery of $\bm{S}^{-1}$. We do \emph{not} claim $\bm{G}\approx\bm{S}$ on
CLIP features; we test the resulting anisotropy--gain prediction empirically
(Section~\ref{sec:experiments}) rather than assuming the model holds.

The rank-$K$ term decomposes as
$\sum_k\pi_k\mu_k\mu_k^\top=\bar\mu\bar\mu^\top+\bm{S}_B$, where
$\bar\mu=\sum_k\pi_k\mu_k$ and
$\bm{S}_B=\sum_k\pi_k(\mu_k-\bar\mu)(\mu_k-\bar\mu)^\top$.
Thus $\bm{G}=\bm{S}+\bar\mu\bar\mu^\top+\bm{S}_B$, and
$(\bm{G}+\gamma\bm{I})^{-1}$ down-weights all three components: the
non-discriminative shared mean \citep{liang2022mind}, high within-class variance in
$\bm{S}$, and discriminative between-class scatter $\bm{S}_B$. CARe therefore
performs a trade-off rather than clean transformation: it helps when suppressing the
shared mean and dominant within-class directions outweighs compression of
$\bm{S}_B$. For $\ell_2$-normalized features, $\mathrm{Tr}(\bm{G})=1$, so the mean
term can remain substantial even when $\mathrm{Tr}(\bm{S})$ is small. Empirical
gains (Section~\ref{sec:experiments}) suggest this trade-off is favorable in
practice, but the theory does not guarantee it.

\begin{remark}[Bayes-optimal discriminant under shared covariance]
\label{rem:bayes-optimal}
Under Assumption~\ref{ass:shared-cov}, maximizing the Gaussian log-posterior yields the
classical linear discriminant \citep{fisher1936use,fukunaga1990introduction,bishop2006pattern}:
\begin{equation}
\hat{y}(f) \;=\; \arg\max_{k\in[K]}\Bigl[\,
\underbrace{f^\top \bm{S}^{-1} \boldsymbol{\mu}_k}_{\text{data term}}
\;-\; \underbrace{\tfrac{1}{2} \boldsymbol{\mu}_k^\top {S}^{-1} \boldsymbol{\mu}_k}_{\text{bias}}
\;+\; \underbrace{\log\pi_k}_{\text{prior}}\Bigr].
\label{eq:bayes-lda}
\end{equation}

The data term $f^\top \bm{S}^{-1} \boldsymbol{\mu}_k$ is a Mahalanobis-weighted inner
product. It is precisely the structure of the  CARe  numerator
(Eq.~\ref{eq:care-score}), which equals $f^\top( \bm{G}_t+\gamma \bm{I})^{-1}w_k$
and reduces as $\gamma\to0^{+}$ to $f^\top \bm{G}_t^{-1}w_k$: under
$w_k\propto \boldsymbol{\mu}_k$, and $\bm{G}$ is
substituted for $ \bm{S}$, the two {numerators} agree up to a positive class-independent scalar. The
full CARe score additionally applies the class-dependent normalization
$\|\bm{M}w_k\|_2$, so it does not in general rank classes identically to LDA; the
correspondence is at the level of the numerator (the decision direction), not the
complete rule. The limit is formal, since $ \bm{G}_t$ is singular for $t<D$ and $\gamma>0$ is what keeps the transform well defined in practice. The cosine rule scores by $f^\top\mu_k$, effectively replacing $\bm{S}^{-1}$ with $\bm{I}$. The data terms $f^\top\bm{S}^{-1}\mu_k$ and $f^\top\mu_k$ induce identical class rankings for all means and test features iff $\bm{S}\propto\bm{I}$. This equivalence concerns only the linear data term; the full rules also differ in cosine normalization $\|\mu_k\|_2$ and Bayes bias and priors.

\end{remark}

\paragraph{What the covariance-aware score inherits from Eq.~\ref{eq:bayes-lda}.}
In its cosine form (Eq.~\ref{eq:care-score}), and using that $\|\tilde{f}\|_2$ is
class-independent, the covariance-aware decision rule is
\begin{equation}
\hat{y}_{\text{CARe}}(f) \;=\; \arg\max_{k\in[K]}
\frac{f^\top( \bm{G}+\gamma {I})^{-1}w_k}{\|\tilde{w}_k\|_2},
\qquad \tilde{w}_k =  \bm{M}w_k .
\label{eq:care-rule}
\end{equation}
Under the idealizations of Remark~\ref{rem:bayes-optimal} ($\gamma\to0^{+}$,
$w_k\propto\boldsymbol{\mu}_k$), the numerator is proportional to the Bayes-optimal
data term, so the two rules share a decision \emph{direction} but not the rule
itself. The Bayes bias $\tfrac{1}{2}\boldsymbol{\mu}_k^\top\bm{S}^{-1}\boldsymbol{\mu}_k$
enters additively, whereas CARe divides by $\|\tilde{w}_k\|_2$; an additive and a
multiplicative penalty on large-norm representatives induce different boundaries.
CARe also carries no $\log\pi_k$ term, as priors are not estimable without labels.
We keep the cosine normalization to preserve scale-invariance and keep CARe a
drop-in replacement, and analyze the decision direction, where the dependence on
anisotropy arises.

This correspondence holds only in the stated direction and under the idealized model. LDA uses the shared covariance $\bm{S}$, whereas CARe uses the total second moment $\bm{G}=\bm{S}+\sum_k\pi_k\mu_k\mu_k^\top$, since $\bm{S}$ is unidentifiable without labels. We do not claim that $\bm{G}^{-1}$ approximates $\bm{S}^{-1}$. Rather, the model isolates the mechanism CARe corrects: directional reweighting that down-weights high-variance directions. Our empirical results (Section~\ref{sec:experiments}) do not rely on this proxy.

\paragraph{Worst-case angular discrepancy under anisotropy.}
\label{sec:theory-cost}
While Remark~\ref{rem:bayes-optimal} establishes qualitative suboptimality,
Theorem~\ref{thm:anisotropy} quantifies the worst-case angular discrepancy between
the Euclidean mean-difference direction $\Delta$ and its covariance-preconditioned
counterpart $\bm{S}^{-1}\Delta$ as a function of the class-covariance condition
number $\kappa$. This result concerns only the angle between two linear directions,
not classification error, decision-boundary displacement, or ranking disagreement.
Moreover, the cosine classifier depends on normalized representatives, while the
LDA discriminant also includes class-dependent norm and prior terms that are not
modeled by the theorem.

\begin{theorem}[Cosine direction error is governed by anisotropy]
\label{thm:anisotropy}
Let $\kappa=\lambda_{\max}( \bm{S})/\lambda_{\min}( \bm{S})$ be the condition number of the class covariance. For any two classes $k\neq k'$ with mean difference $ {\Delta}= \boldsymbol{\mu}_k- \boldsymbol{\mu}_{k'}\neq {0}$, let $\theta$ be the angle between the cosine discriminant direction $ {\Delta}$ and the Bayes-optimal direction $ \bm{S}^{-1} {\Delta}$. Then
\begin{equation}
\cos\theta \;=\; \frac{ {\Delta}^\top \bm{S}^{-1} {\Delta}}{\| {\Delta}\|_2\,\| \bm{S}^{-1} {\Delta}\|_2},
\qquad
\inf_{ {\Delta}\neq {0}}\cos\theta \;=\; \frac{2\sqrt{\kappa}}{1+\kappa}.
\label{eq:kantorovich-bound}
\end{equation}

The worst-case alignment decreases monotonically from $1$ (isotropic, $\kappa=1$) toward $0$ as $\kappa\to\infty$; equivalently, the maximal angular error $\arccos\!\bigl(2\sqrt{\kappa}/(1+\kappa)\bigr)$ increases with anisotropy. This bound is tight: equality is attained by a specific vector within the span of the extreme eigenvectors of $\bm{S}$ (specifically, a balanced linear combination of the eigenvectors corresponding to $\lambda_{\max}$ and $\lambda_{\min}$.
\end{theorem}

Proof of this Theorem is given in Appendix~\ref{app:proofs}. Theorem~\ref{thm:anisotropy} bounds the angular gap between the cosine
direction $\Delta$ and the \emph{Bayes-optimal} direction
$\bm{S}^{-1}\Delta$. It therefore provides an upper bound on the directional
gain achievable by any covariance-aware correction. CARe is a regularized,
label-free instance of such a correction, using the accumulated $\bm{G}$ in
place of $\bm{S}$ and a finite $\gamma$. CARe attains this bound only in the
idealized limit $\gamma\to 0^{+}$ with $\bm{G}=\bm{S}$; for finite $\gamma$, it
can close the gap only partially (Remark~\ref{rem:geometry}). The theorem
therefore serves as the theoretical counterpart to our central empirical
observation: the potential for mis-ranking increases with anisotropy. It does
not guarantee a corresponding realized gain for CARe, which we measure
directly in Section~\ref{sec:experiments}.

\paragraph{Stability under online estimation.}
\label{sec:theory-stability}
The transform $ \bm{M}$ is formed from a second moment accumulated over a finite
stream prefix, and its inverse square root most amplifies the discrepancies in the
low-variance directions, which are the least reliably determined from a few samples. Let $\widehat{\bm{G}}$ and $\bm{G}$ be same-scale second-moment matrices such that $\widehat{\bm{G}}$
 is a perturbation of $\bm{G}$ within budget $\varepsilon$. The Proposition~\ref{prop:stability} gives a static perturbation bound: if $\|\widehat{\bm{G}}-\bm{G}\|\le\varepsilon$, the induced metrics differ by at most $\varepsilon\gamma^{-2}$. It characterizes stability in under-sampled directions during warm-up, not
convergence as $t\to\infty$; the accumulated $\widehat{\bm{G}}$ grows with $t$
rather than converging to a fixed matrix.
\begin{proposition}[$\gamma$ bounds estimation-error amplification]
\label{prop:stability}
For any $\widehat{ \bm{G}}\succeq {0}$ the transform satisfies
$\|(\widehat{ \bm{G}}+\gamma \bm{I})^{-1/2}\|_2\le\gamma^{-1/2}$. If moreover
$\|\widehat{ \bm{G}}- \bm{G}\|_2\le\varepsilon$, the induced metrics
$\widehat{ \bm{A}}=(\widehat{ \bm{G}}+\gamma {I})^{-1}$ and
$ \bm{A}=( \bm{G}+\gamma \bm{I})^{-1}$ obey
\begin{equation}
\|\widehat{ \bm{A}}- \bm{A}\|_2 \;\le\; \varepsilon\,\gamma^{-2}.
\end{equation}
\end{proposition}

Proof is given in Appendix~\ref{app:proofs}. The bound makes the role of $\gamma$ explicit: it
trades fidelity to the exact $\gamma=0$ geometry, which degrades as $\gamma$ grows,
against sensitivity to estimation error, which decays as $\gamma^{-2}$. The base rule
sits at the $\gamma\to\infty$ endpoint, where the metric is isotropic and carries no
covariance information at all. Separately, as the spectrum of $\bm{G}_t$ grows with $t$, a fixed $\gamma$ becomes
progressively less influential (the accumulation argument of
Section~\ref{sec:care}); this governs late-stream behaviour, whereas
Proposition~\ref{prop:stability} concerns fixed-$t$ stability. This predicts that  CARe  should be insensitive to the
precise value of $\gamma$ once the stream is underway.
Section~\ref{sec:exp-ablation} confirms this: accuracy varies by less than a point
across four orders of magnitude of $\gamma$, and we fix a single value for all datasets
and both backbones.

\section{Experiments and Results}\label{sec:experiments}
\label{sec:exp-setup}
We use CLIP~\citep{radford2021learning} with ViT-B/16 and ResNet-50 as frozen
backbones. All experiments run on a single A100 GPU (40~GB). The datasets details, algorithm and the baseline are provided in
Appendix~\ref{app:dataset_and_baselines}. CARe introduces no
trainable parameters or additional forward passes, and reuses the second-moment
matrix of the test features. It adds only one hyperparameter beyond those inherited
from the base adapter, namely the covariance regularizer $\gamma$ in
Eq.~\ref{eq:transform}. We retain \citep{guan2026statistics} settings for the ridge parameter
$\lambda_{\text{ridge}}$ and pseudo-label threshold $\tau_p$, and fix
$\gamma=10$ across all datasets and backbones without per-dataset tuning. The
transform is refreshed every $\eta$ steps. This is an efficiency setting rather
than a tuned hyperparameter, since accuracy is unchanged for
$\eta\in[1,100]$ (Appendix~\ref{app:cost}); we therefore use $\eta=1$ for all
reported results. We evaluate standard OOD and fine-grained cross-domain image
classification benchmarks against recent TTA baselines reproduced from their
released code. The reported accuracies of CARe are averaged over up to five random stream seeds to account for the order sensitivity of the online stream (Full sensitivity details are provided in Appendix. \ref{app:seeds}). 

\paragraph{Performance on OOD Generalization.}
Table~\ref{tab:ood} reports top-1 accuracy on ImageNet and four of its OOD variants for both backbones.  CARe  achieves the best average accuracy with both backbones, reaching 51.80\% with RN-50 and 67.39\% with ViT-B/16, as well as the highest OOD average of 48.63\% and 66.02\%, respectively. Relative to SCA, its closest methodological baseline, CARe improves average accuracy by +1.41\% and +1.97\%, and OOD accuracy by +1.27\% and +1.63\%, on RN-50 and ViT-B/16, respectively. These gains require only replacing the cosine score with the covariance-aware score, with no changes to the adaptation procedure or per-dataset tuning.  CARe  also outperforms recent competing baselines showing that the covariance correction consistently improves OOD recognition across different architectures.

\begin{table}[ht!]
\centering
\caption{OOD generalization on RN-50 and CLIP ViT-B/16. {Bold} marks the best result.}
\label{tab:ood}
\vspace*{-2mm}
\resizebox{0.9\textwidth}{!}{%
\begin{tabular}{cl ccccc cc}
\toprule
 \textit{Arch.} &\textit{Method} & \textit{ImageNet} & \textit{ImageNet-A} &
\textit{ImageNet-V2} & \textit{ImageNet-R} & \textit{ImageNet-S}
& \textit{Average} & \textit{OOD Avg.} \\

\midrule

\multirow{6}{0.2em}{\rotatebox{90}{RN-50}}
& TDA & 61.13 & 30.64 & 54.90 & 62.03 & 37.90 & 49.32 & 46.37 \\

& DPE & 63.07 & 29.61 & \textbf{57.05} & \textbf{64.41} & 39.37 & 50.70 & 47.61 \\

& SCA & 62.49 & 31.01 & 55.74 & 63.11 & 39.59 & 50.39 & 47.36 \\

& ADAPT & 62.72 & 32.44 & 55.19 & 62.94 & \bf 40.91 & 50.84 & 47.87 \\

& PTA & 63.66 & 32.39 & 53.55 & 60.46 & 39.39 & 49.89 & 46.45 \\

\rowcolor{green!12}
& {CARe (Ours)} & \bf 64.48 & \bf 34.52 & 56.17 &
63.29 &  40.55 & \bf 51.80 & \bf 48.63 \\

\midrule

\multirow{6}{0.2em}{\rotatebox{90}{ViT/B-16}}
& TDA & 70.09 & 60.41 & 65.00 & 80.86 & 49.97 & 65.27 & 64.06 \\

& DPE & 71.39 & 59.33 & \bf 65.80 & 79.79 & 51.92 & 65.65 & 64.21 \\

& SCA & 69.53 & 61.49 & 63.26 & 81.26 & 51.55 & 65.42 & 64.39 \\

& ADAPT & 70.84 & 60.52 & 65.09 & \bf 81.59 & 51.91 & 65.99 & 64.78 \\

& PTA & 69.77 & 60.50 & 64.56 & 80.11 & 51.56 & 65.30 & 64.18 \\

\rowcolor{green!12}
& {CARe (Ours)} & \bf 72.87 & \bf 64.37 & 65.47 &
81.04 & \bf 53.20 & \bf 67.39 & \bf 66.02 \\

\bottomrule
\end{tabular}}
\end{table}


\begin{table}[ht!]
\centering
\caption{Cross-domain generalization performance on RN-50 and ViT-B/16. {Bold} marks the best result in each column.}
\label{tab:crossdomain}
\vspace*{-2mm}
\setlength{\tabcolsep}{4.5pt}
\renewcommand{\arraystretch}{1.1}
\resizebox{0.9\textwidth}{!}{%
\begin{tabular}{cl cccccccccc c}
\toprule
\textit{Arch.} & \textit{Method} & \textit{Flowers} & \textit{DTD} & \textit{Pets} & \textit{Cars}
 & \textit{UCF101} & \textit{Caltech} & \textit{Food101} & \textit{SUN397}
 & \textit{Aircraft} & \textit{EuroSAT} & \textit{Average} \\
\midrule
\multirow{6}{0.2em}{\rotatebox{90}{RN-50}}
&TDA   & 62.14 & 41.37 & 84.91 & 57.91 & 60.36
& 87.64 & 74.21 & 62.01 & 18.16 & 27.71 & 57.64 \\
&DPE            & 68.22 & 49.63 & 86.51 & 59.84 & 61.41
& 90.27 & 78.01 & 63.67 & 20.46 & 42.31 & 62.07 \\
&SCA   & 70.24 & 52.25 & 85.75 & 59.57 &  {65.87}
&  \textbf{89.78} & 76.12 & 65.08 & 19.77 & 41.01 & 62.54 \\
&ADAPT & 70.63 & 51.21 & 86.94 & 57.83 & 62.57
& 88.81 & 76.18 & 65.66 & 18.54 & 49.82 & 62.82 \\
&PTA   & 67.91 & 45.62 & 87.31 & 58.81 & 65.87
& 89.19 & 76.24 & 63.47 & 18.06 &  \textbf{52.21} & 62.47 \\
\rowcolor{green!12}
& {CARe (Ours)} &  \textbf{71.62} &  \textbf{55.14} &  \textbf{87.65} &  \textbf{61.27}
& \textbf{67.12} & {89.25} &  \textbf{78.17} &  \textbf{66.32} &  \textbf{21.00} & {49.73} &  \textbf{64.66} \\
\midrule
\multirow{6}{0.2em}{\rotatebox{90}{ViT-B/16}}
&TDA   & 72.13 & 44.51 & 89.94 & 67.72 & 70.24
& 94.62 & 85.71 & 68.38 & 24.84 & 61.94 & 68.00 \\
&DPE            & 74.48 & 54.83 & 90.61 & 66.77 & 71.03
& 95.32 & 85.56 & 70.64 &  {29.53} & 55.14 & 69.39 \\
&SCA   & 75.72 & 55.32 & 91.08 & 67.50 & 72.56
& 94.69 & {85.55} & 68.19 & 28.80 & 55.96 & 69.54 \\

&ADAPT & 72.69 & 47.20 & 91.58 & 67.91 & 69.96
& 94.02 & 85.14 & 68.51 & 26.18 &  \textbf{62.37} & 68.56 \\
&PTA   & 75.91 & 47.02 & 90.49 & 69.13 & 72.54
&  \textbf{95.33} & 85.92 & 69.76 & 26.76 & 60.94 & 69.38 \\
\rowcolor{green!12}
& {CARe (Ours)} &  \textbf{77.87} &  \textbf{56.97} &  \textbf{92.48} &  \textbf{69.67}
&  \textbf{74.02} & 94.16 &  \textbf{86.80} &  \textbf{71.33} & \textbf{29.22} & 62.00 &  \textbf{71.45} \\
\bottomrule
\end{tabular}%
}
\end{table}

\paragraph{Performance on Cross-Domain Generalization.}
Table~\ref{tab:crossdomain} evaluates  CARe  on ten fine-grained cross-domain recognition datasets spanning textures, scenes, aircraft, satellite imagery, and other domains.  CARe  achieves the highest average accuracy with both backbones, reaching 64.66\% with RN-50 and 71.45\% with ViT-B/16, improving over SCA, by +1.91\% and +2.12\%, respectively, while also outperforming the strongest baselines such as ADAPT, DPE and PTA by a noticeable margin.

\subsection{Analysis and ablations}
\label{sec:exp-ablation}

\begin{figure}[t]
  \centering
  \includegraphics[width=0.92\linewidth]{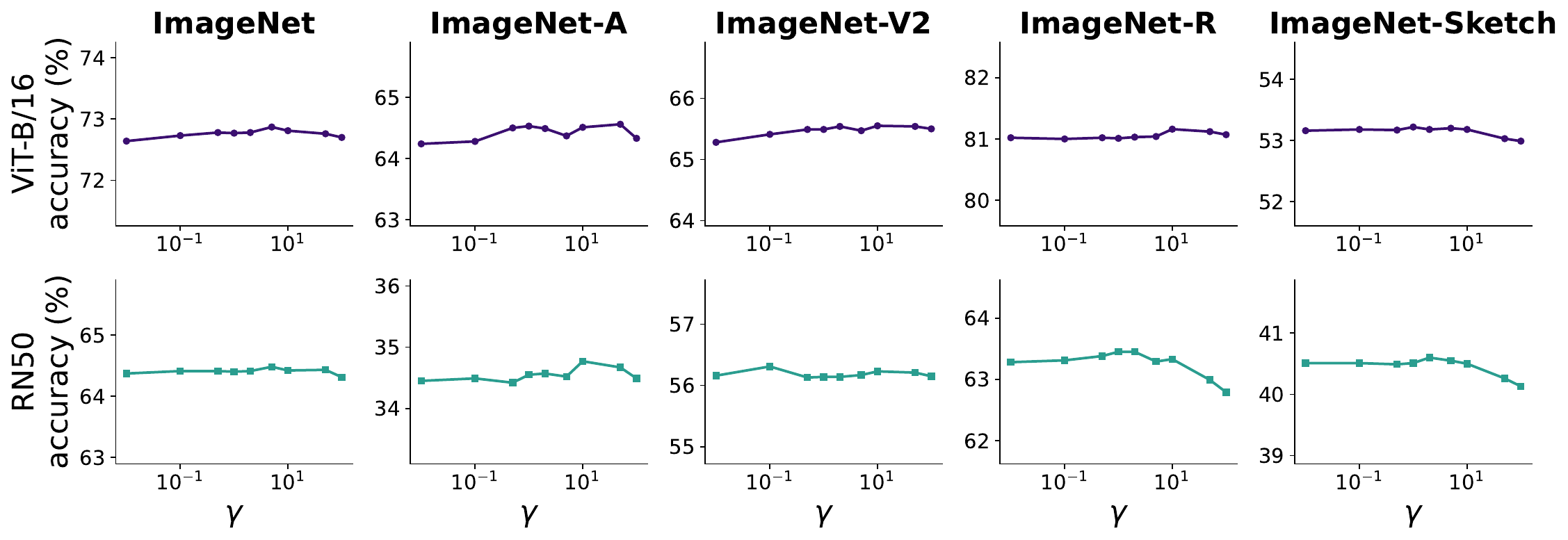}
  \caption{$\gamma$ sweep over three orders of magnitude on the ImageNet-OOD variants
  for both backbones. Accuracy varies by less than a point across the entire range on
  every dataset and backbone, with a mild decline at the largest $\gamma$ on the
  shorter streams.}
  \label{fig:gamma_vs_acc}
\end{figure}

\paragraph{Robustness to $\gamma$.}
Figure~\ref{fig:gamma_vs_acc} sweeps $\gamma$ across three orders of magnitude on ImageNet-OOD variants for both backbones. Accuracy varies by less than one point throughout, as the growing spectrum of $\bm{G}_t$ quickly dominates $\gamma$; Proposition~\ref{prop:stability} bounds sensitivity during the early stream when regularization is non-negligible. The mild degradation at large $\gamma$ is strongest on smaller datasets. We therefore use a single $\gamma$ across all datasets and backbones, without per-dataset tuning.

\paragraph{Anisotropy predicts the gains of CARe.}
\begin{wrapfigure}[13]{r}{0.6\linewidth}
    \centering
    \begin{subfigure}[t]{0.48\linewidth}
        \centering
        \includegraphics[width=0.95\linewidth]{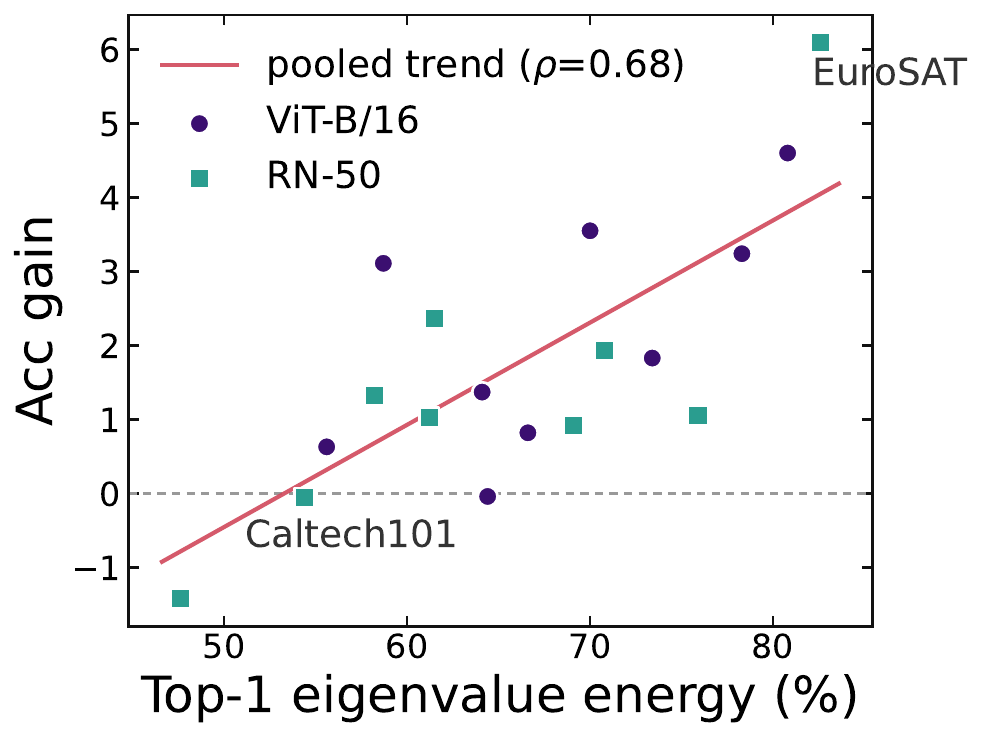}
        \caption{}
        \label{fig:aniso_gain}
    \end{subfigure}
    \begin{subfigure}[t]{0.48\linewidth}
        \centering
        \includegraphics[width=0.95\linewidth]{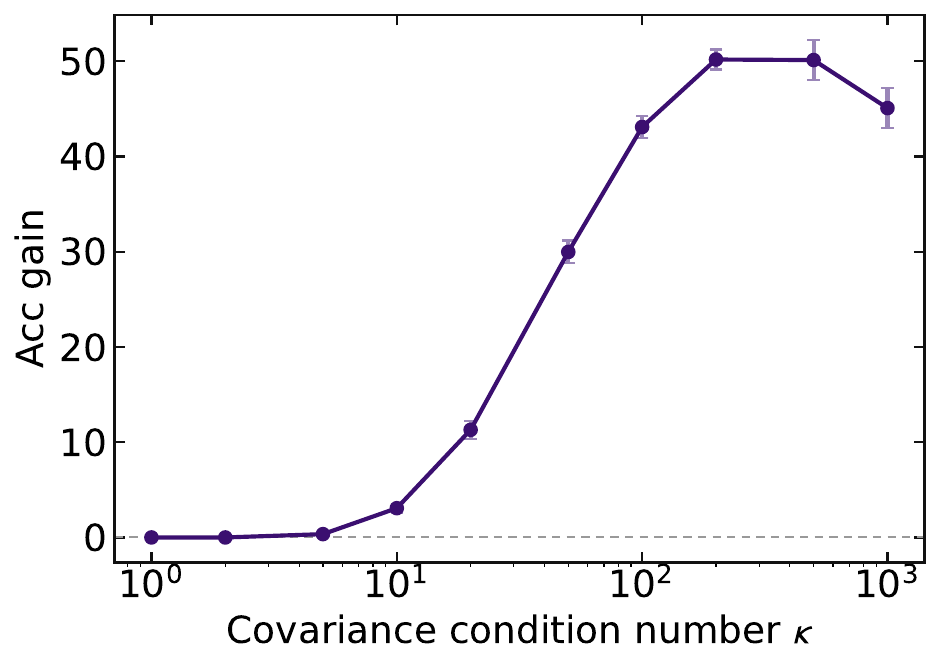}
        \caption{}
        \label{fig:kappa_gain}
    \end{subfigure}
\vspace*{-2mm}
    \caption{\small{Accuracy gains of $s_k^{CARe}$ over $s_k^{base}$. (a) Cross-domain datasets. (b) Synthetic Gaussian data}}
    \label{fig:comparison}
\end{wrapfigure}
\vspace*{-2mm}
Theorem~\ref{thm:anisotropy} shows that a transform-based correction has more room to help when $\bm{S}$ is
anisotropic. We evaluate this prediction on real and synthetic data in Figures~\ref{fig:aniso_gain} and~\ref{fig:kappa_gain}. On the cross-domain benchmarks, we use the fraction of variance explained by the leading covariance direction as a proxy for spectral concentration, since the condition number $\kappa$ is not directly measurable in this setting. The general trend in accuracy gains is positively associated with this proxy for both backbones (RN-50: Spearman's $\rho=0.70$, $p<0.05$; ViT-B/16: $\rho=0.67$, $p=0.05$). We also found that, the gain across relatively isotropic datasets such as Caltech101 is lower than that of highly anisotropic datasets such as EuroSAT. To isolate this relationship, we generate synthetic Gaussian features with fixed class geometry and a shared covariance whose condition number $\kappa$ varies. The gain is zero at $\kappa=1$, increases with anisotropy, and saturates before declining at very large $\kappa$. The initial trend agrees with Theorem~\ref{thm:anisotropy}, while the decline is consistent with Proposition~\ref{prop:stability}, as finite-sample covariance estimation becomes unreliable in low-variance directions. Together, these results support anisotropy as a key driver of CARe's benefit, while also showing that the realized gain depends on covariance--class alignment and estimation stability.

\paragraph{Factor ablation: transforming $f$, $\bm{W}$ or both.}
We ablate the components of {CARe} at $\gamma=10$ across the ten cross-domain
benchmarks and both backbones (Table~\ref{tab:factor}), applying the transform
$\bm{M}$ to the test feature $f$ only, the class representatives $\bm{W}$ only, or
both, while keeping all other components fixed. Transforming $\bm{W}$ alone has
little effect and can slightly degrade accuracy (e.g.\ $62.54\to62.23$ on RN-50),
whereas transforming $f$ alone yields modest gains ($+0.62$ and $+0.84$ points on
RN-50 and ViT-B/16). Applying $\bm{M}$ to both sides produces the largest gains
($+2.12$ and $+1.91$) and the best mean accuracy overall. The full ordering
(${\text{None} < \bm{W}\text{only} < f \text{only} < \text{Both}}$) holds strictly on
ViT-B/16; on RN-50 the same ordering holds except that $\bm{W}$ only falls marginally
below None. Only the two-sided transform realizes the full Mahalanobis inner
product $f^\top(\bm{G}+\gamma\bm{I})^{-1}w_k$ (Remark~\ref{rem:geometry}); the
one-sided variants capture only part of the covariance correction.

\begin{table}[ht!]
\centering
\caption{Factor ablation on the cross-domain benchmarks: applying the covariance-aware transform $\bm{M}$ to the feature ($f$), the class
representatives ($\bm{W}$), or both.}
\label{tab:factor}
\footnotesize
\vspace*{-2mm}
\setlength{\tabcolsep}{3.5pt}
\resizebox{0.8\textwidth}{!}{%
\begin{tabular}{llccccccccccc}
\toprule
\textit{Arch.} & \textit{Transform} & \textit{Flowers} & \textit{DTD} & \textit{Pets} & \textit{Cars} & \textit{UCF101} & \textit{Caltech} & \textit{Food101} & \textit{SUN397} & \textit{Aircraft} & \textit{EuroSAT} & \textit{Mean} \\
\midrule
\multirow{4}{*}{\rotatebox{90}{RN-50}}
 & None       & 70.24 & 52.25 & 85.75 & 59.57 & 65.87 & \textbf{89.78} & 76.12 & {65.08} & 19.77 & 41.01 & 62.54 \\
 & $\bm{W}$ only   & 68.57 & 52.19 & 84.66 & 58.38 & 65.53 & 89.90 & 76.89 & 64.87 & 18.75 & 42.59 & 62.23 \\
 & $f$ only   & 69.55 & 52.78 & 87.14 & 60.55 & 66.14 & 89.29 & 76.28 & 65.24 & \textbf{21.27} & 43.33 & 63.16 \\
 & Both (CARe) & \textbf{71.62} & \textbf{55.14} & \textbf{87.65} & \textbf{61.27} & \textbf{67.12} & 89.25 & \textbf{78.17} & \textbf{66.32} & 21.00 & \textbf{49.73} & \textbf{64.66} \\
\midrule
\multirow{4}{*}{\rotatebox{90}{ViT-B/16}}
 & None       & 75.72 & 55.32 & 91.08 & 67.50 & 72.56 & \textbf{94.69} & 85.55 & 68.19 & 28.80 & 55.96 & 69.54 \\
 & $\bm{W}$ only   & 75.36 & 54.61 & 90.91 & 67.31 & 72.11 & 94.56 & 86.09 & 68.76 & 26.82 & 59.36 & 69.58 \\
 & $f$ only   & 75.88 & 55.73 & 92.18 & 68.83 & 73.01 & 93.63 & 86.41 & 69.67 & \textbf{29.79} & 58.69 & 70.38 \\
 & Both (CARe) & \textbf{77.87} & \textbf{56.97} & \textbf{92.48} & \textbf{69.67} & \textbf{74.02} & 94.16 & \textbf{86.80} & \textbf{71.33} & 29.22 & \textbf{62.00} & \textbf{71.45} \\
\bottomrule
\end{tabular}}
\end{table}

\begin{wrapfigure}[10]{r}{0.4\linewidth}
  \centering
  \includegraphics[width=\linewidth]{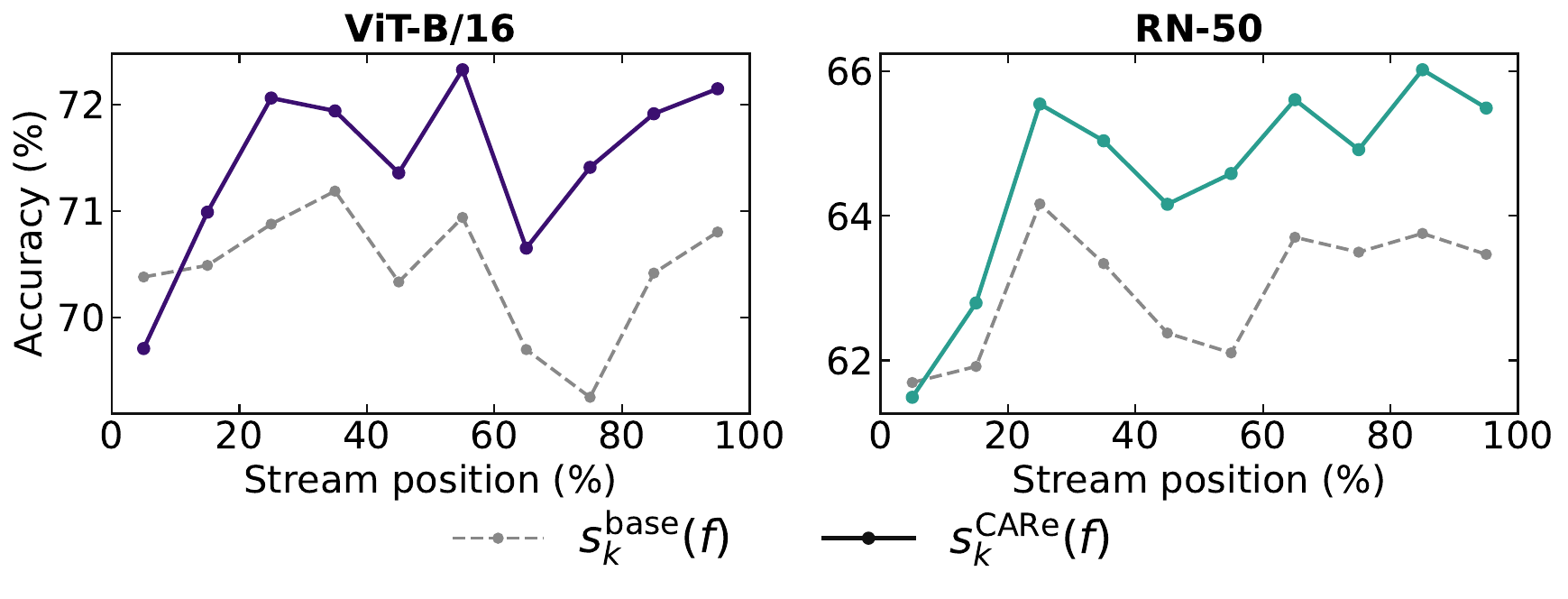}
  \caption{ \small Mean accuracy in successive windows along the stream, averaged across the cross-domain datasets, for the base rule and CARe on ViT-B/16 (left) and RN-50
  (right).}
  \label{fig:streamlen}
\end{wrapfigure}

\paragraph{Behaviour along the online stream.}
Because $\bm{G}$ is accumulated online, CARe's early-stream behavior depends on how quickly its estimate becomes informative. Figure~\ref{fig:streamlen} shows that, with few samples, $\gamma$ dominates the spectrum, making $\bm{M}$ near-isotropic and CARe closely tracks the base rule (Proposition~\ref{prop:stability}). The correction activates within the first 20--30\% of the stream and thereafter maintains a consistent margin. This stable gain is consistent with the $O(1/t)$ decay of the effective regularization (Section~\ref{sec:method}), which makes $\gamma$ progressively negligible as the spectrum accumulates. 
\vspace*{-5mm}

\section{Conclusion and Limitation}
We introduced CARe, a covariance-aware scoring rule for CLIP-based TTA. Our key finding is that existing TTA methods often improve feature or class representations while retaining covariance-agnostic cosine similarity, which can be limiting in anisotropic feature spaces. CARe corrects this mismatch by replacing cosine geometry with a Mahalanobis geometry induced by covariance of online test feature stream. Our theory characterizes when this correction is beneficial and shows that its potential benefit increases with anisotropy. Experiments on real and synthetic benchmarks validate these predictions, demonstrating consistent accuracy gains beating the baseline TTA methods.

We identify two main limitations of CARe. First, it corrects the decision geometry
using the total second moment $\bm{G}$, whereas the optimality analysis concerns
the shared covariance $\bm{S}$. These coincide only when the rank-$K$ mean term is
negligible, so the theory motivates the mechanism rather than predicting realized
gains. Second, improvements are mostly concentrated on anisotropic
datasets, with limited effect under near-isotropy.

\subsubsection*{AI Use Statement}
The authors used GenAI tools exclusively for language editing, including improving the readability and grammatical clarity of the manuscript, and for assisting with the refactoring of source code for final submission. All AI-assisted outputs were critically reviewed, revised, and verified by the authors to ensure their accuracy, integrity, and originality. The authors retain full responsibility for the content and conclusions of this work.

\subsubsection*{Ethics Statement}
This study did not involve human participants, animal subjects, or any associated personal or identifiable data. All data processing and computational analyses were performed in accordance with applicable institutional policies and ethical guidelines.

\subsubsection*{Reproducibility Statement}
To facilitate the reproducibility of the reported results, the custom source code that includes TTA evaluation script and details to the usage of publicly available datasets have been provided as supplementary materials with the manuscript submission. Subject to any applicable repository or licensing requirements, these materials will be made publicly available upon acceptance of the paper.

\bibliography{iclr2026_conference}
\bibliographystyle{iclr2026_conference}

\appendix

\section{Related Works}
\label{app:related}

\subsection{Zero-shot Vision--Language Models}

Large-scale VLMs such as CLIP \citep{radford2021learning} align images and text in a
shared embedding space through contrastive pretraining on web-scale data. This
alignment enables zero-shot classification: an image is labelled by its nearest
class-name embedding, without any task-specific training. The resulting classifier is
attractive for deployment because it requires no labelled target data and no gradient
updates, but its accuracy degrades under distribution shift between the pretraining and
deployment domains \citep{sheng2025r, shu2022test}. A substantial line of work adapts
VLMs to downstream tasks through prompt learning \citep{zang2022unified,
zhou2022learning, zhou2022conditional, khattak2023maple} or adapter modules
\citep{gao2024clip, zhang2021tip}, but these methods assume access to labelled target
data and offline training.

A property of these classifiers that is central to our work is the geometry of the
learned embedding space. Contrastive pretraining does not produce isotropic
representations: \citep{liang2022mind} show that image and text embeddings occupy
narrow, well-separated cones, so image features share a large common mean direction and
their second moment is dominated by a small number of directions. Our diagnostic in
Figure~\ref{fig:anisotropy} confirms this at deployment time across both
in-distribution and shifted benchmarks. Existing adaptation methods inherit this
geometry unchanged, since they modify the class representatives but continue to compare
them against test features in the ambient space. Our contribution is to correct the
comparison itself rather than the representatives.

\subsection{Online Test-Time Adaptation}

TTA adapts a model to a shifting target distribution using only unlabelled test data as
it arrives \citep{wang2020tent, sun2020test}. Gradient-based methods update model
parameters online by minimizing a surrogate objective such as prediction entropy
\citep{wang2020tent, niu2023towards}, optionally with sample filtering or
regularization to curb error accumulation \citep{niu2022efficient}; continual variants
further address error drift over long non-stationary streams \citep{wang2022continual}.
For vision--language models, a complementary and increasingly popular family avoids
backpropagation entirely, instead maintaining a lightweight auxiliary state over the
test stream. TDA \citep{karmanov2024efficient} keeps a small key--value cache of
confident samples and augments the zero-shot logits with a cache-based term; subsequent
methods refine the memory with dual static/dynamic stores
\citep{zhang2024dual}, regional bootstrapping \citep{zhang2024boostadapter}, evolving
prototypes \citep{zhang2024dualpe}, continuous distributional estimates
\citep{han2026dota}, or accumulated first- and second-order feature statistics
\citep{guan2026statistics}.

\subsection{Second-Order Statistics and Discriminant Geometry}

Using second-order feature statistics to shape a decision rule is classical: linear
discriminant analysis \citep{fisher1936use, fukunaga1990introduction, bishop2006pattern}
replaces the Euclidean inner product with a Mahalanobis one induced by the shared
covariance, and the Mahalanobis distance itself dates to \citep{mahalanobis1936generalized}.
Whitening transforms built from feature covariance are also standard in representation
learning and domain adaptation as a means of removing dominant, non-discriminative
directions.

Within backpropagation-free VLM adaptation, two recent methods make use of second-order
statistics. SCA \citep{guan2026statistics}, the adapter we build on, accumulates first-
and second-order feature statistics and solves a regularised linear system to obtain
its class representatives, but scores those representatives with the same ambient-space
readout as the rest of the family. ADAPT \citep{zhang2026backpropagation} performs
probabilistic Gaussian alignment, estimating per-class means and a centered shared
covariance from knowledge banks and classifying under a full discriminant model. CARe
is distinct from both. Unlike SCA, we alter the readout rather than the construction of
the representatives; unlike ADAPT, we do not estimate class-conditional statistics or
build a generative classifier. We use only the uncentred total second moment the adapter already accumulates and
apply it as a metric on the representatives that adapter produces, reducing
exactly to the base rule when the feature distribution is isotropic.
\section{Experimental Configurations}
\label{app:dataset_and_baselines}
\subsection{Datasets.}
Following the standard TTA protocol for VLMs~\citep{sheng2025r,karmanov2024efficient}, we evaluate on two benchmark suites.
The natural distribution shift suite comprises
ImageNet~\citep{deng2009imagenet} and its four OOD variants including ImageNet-A~\citep{hendrycks2019nae},
ImageNet-V2~\citep{recht2019imagenet}, ImageNet-R~\citep{hendrycks2020many},
and ImageNet-Sketch~\citep{wang2019learning} which share ImageNet's label space
under increasing shift. The fine-grained cross-domain suite comprises ten
datasets spanning diverse recognition tasks:
Aircraft~\citep{maji2013fine}, Caltech101~\citep{fei2004learning},
Cars~\citep{krause20133d}, DTD~\citep{cimpoi2014describing},
EuroSAT~\citep{helber2019eurosat}, Flowers102~\citep{nilsback2008automated},
Food101~\citep{bossard2014food}, Pets~\citep{parkhi2012cats},
SUN397~\citep{xiao2010sun}, and UCF101~\citep{soomro2012ucf101}.

\subsection{Baselines.}
We compare CARe against the backpropagation-free methods, such as:
TDA~\citep{karmanov2024efficient}, DPE~\citep{zhang2024dualpe},
PTA~\citep{huang2026prototype}, ADAPT \citep{zhang2026backpropagation} and SCA~\citep{guan2026statistics}. All these methods
were reproduced using the officially released code under the unified prompt template "a photo of a $<$class$>$".

\subsection{Algorithm}
The complete TTA algorithm for CARe is provided in Algorithm~\ref{alg:care}. The default settings are as follows: $\lambda_\text{ridge}=10^4$ (ViT-B/16) or $10^5$ (RN-50); $\gamma=10$;
$\tau_p=0.2$; $\eta=1$.

\begin{algorithm}[ht!]
\caption{CARe: covariance-aware online TTA (one stream)}
\label{alg:care}
\begin{algorithmic}[1]
\Require frozen CLIP encoders $\phi_I,\phi_T$; text classifier
$\bm{W}^{\text{Z.S}}=[w^{\text{Z.S}}_1,\dots,w^{\text{Z.S}}_K]$; ridge
$\lambda_{\text{ridge}}$; metric regularizer $\gamma$; pseudo-label threshold
$\tau_p$; refresh interval $\eta$
\State $\bm{G}_0\gets\bm{0}$,\quad $\bm{P}_0\gets\bm{0}$,\quad $\bm{M}\gets\text{undefined}$
\For{$t=1,2,\dots$ (test samples arrive online)}
  \State $f_t \gets \phi_I(x_t)/\lVert\phi_I(x_t)\rVert_2$
  \State compute zero-shot score $s_k^{\text{Z.S}}$ and posterior $p_k^{\text{Z.S}}$
         \Comment{Eq.~\ref{eq:zeroshot}}
  \Statex \textit{// Adaptive soft pseudo-label}
  \State $\delta\gets p_{\max}+\tau_p\,(1-p_{\max})$, \; $p_{\max}=\max_k p^{\text{Z.S}}_k$
  \State $\mathcal{C}\gets$ smallest set of top-ranked classes with
         $\sum_{k\in\mathcal{C}} p^{\text{Z.S}}_k \ge \delta$
  \State $q_{t,k}\gets p^{\text{Z.S}}_k / \!\!\sum_{j\in\mathcal{C}} p^{\text{Z.S}}_j$
         for $k\in\mathcal{C}$, else $0$
    \Comment{posterior renormalized over $\mathcal{C}$}

  \Statex \textit{// Online accumulation (current sample included before scoring)}
  \State $\bm{G}_t\gets\bm{G}_{t-1}+f_t f_t^\top$,\quad
         $\bm{P}_t\gets\bm{P}_{t-1}+q_t f_t^\top$
  \State $\bm{W}_t\gets(\bm{G}_t+\lambda_{\text{ridge}}\bm{I})^{-1}\bm{P}_t^\top$
    \Comment{Eq.~\ref{eq:sca-w}}
  \Statex \textit{// Covariance-aware transform, refreshed every $\eta$ steps}
  \If{$\bm{M}$ undefined \textbf{or} $t \bmod \eta = 0$}
     \State $\bm{M}\gets(\bm{G}_t+\gamma\bm{I})^{-1/2}$ \Comment{Eq.~\ref{eq:transform}}
  \EndIf
  \State $s^{\text{CARe}}_k\gets\cos(\bm{M}f_t,\,\bm{M}w_{t,k})$, \; $k=1,\dots,K$
    \Comment{Eq.~\ref{eq:care-score}}
  \Statex \textit{// Entropy-based score fusion}
  \State $\hat{y}_t\gets\arg\max_k\big[(1-\alpha)\, s^{\text{Z.S}}_k+\alpha\,s^{\text{CARe}}_k\big]$,
         \; $\alpha$ from predictive entropies \Comment{Eq.~\ref{eq:fusion}}
\EndFor
\end{algorithmic}
\end{algorithm}

\subsection{Synthetic Study Configuration}
\label{app:synthetic}
For Figure~\ref{fig:kappa_gain}, features follow the shared-covariance model
$x\mid y=k\sim\mathcal{N}(\mu_k,\bm{\Sigma})$ with $D=64$, $K=20$, and
$N_{\text{train}}=40$ and $N_{\text{test}}=200$ samples per class. Class means are
drawn i.i.d. as $\mu_k\sim\mathcal{N}(0,\bm{I})$ and held fixed across the sweep.
The shared covariance has condition number exactly $\kappa$, with eigenvalues
log-spaced in $[1,\kappa]$ and a random orthonormal eigenbasis sampled per trial;
thus, anisotropy is the only varying factor. We sweep
$\kappa\in\{1,\ldots,1000\}$ and average over 30 random configurations per point.

\section{Proof of Theoretical Statements}
\label{app:proofs}
\begin{proof}[Proof of Theorem~\ref{thm:anisotropy}]
Let $ \bm{A}= \bm{S}^{-1}\succ {0}$. The eigenvalues of $ \bm{A}$ are the reciprocals of those of $ \bm{S}$, so $ \bm{A}$ has condition number $\kappa_{ \bm{A}}=\lambda_{\max}( \bm{A})/\lambda_{\min}( \bm{A})=\lambda_{\max}( \bm{S})/\lambda_{\min}( \bm{S})=\kappa$. By the \emph{Kantorovich inequality} \citep{kantorovich1948functional}, for any symmetric positive-definite matrix with spectrum in $[\lambda_{\min},\lambda_{\max}]$,
\begin{equation}
\inf_{ {v}\neq {0}}\frac{ {v}^\top \bm{A} {v}}{\| {v}\|_2\,\| \bm{A} {v}\|_2}
\;=\;
\frac{2\sqrt{\lambda_{\max}\lambda_{\min}}}{\lambda_{\max}+\lambda_{\min}}
\;=\;
\frac{2\sqrt{\kappa_{ \bm{A}}}}{1+\kappa_{ \bm{A}}}.
\end{equation}
Substituting $\kappa_{ \bm{A}}=\kappa$ and noting that the left-hand side is exactly $\inf_{ {\Delta}\neq {0}}\cos\theta$ yields Eq.~\ref{eq:kantorovich-bound}. Equality holds when $ {\Delta}$ is in the span of the eigenvectors corresponding to $\lambda_{\min}( \bm{S})$ and $\lambda_{\max}( \bm{S})$ \citep[Sec.~7.4.12]{horn2012matrix}.
\end{proof}

\begin{proof}[Proof of Proposition~\ref{prop:stability}]
Since $\widehat{\bm{G}}\succeq {0}$, the eigenvalues of $\widehat{\bm{G}}+\gamma I$ are at least $\gamma$. Therefore:
\begin{equation}
\|(\widehat{\bm{G}}+\gamma \bm{I})^{-1/2}\|_2 
= \lambda_{\max}\left((\widehat{\bm{G}}+\gamma \bm{I})^{-1/2}\right) 
= \frac{1}{\sqrt{\lambda_{\min}(\widehat{\bm{G}}+\gamma \bm{I})}} 
\le \gamma^{-1/2}.
\end{equation}

Let $\widehat{ \bm{A}}=\widehat{\bm{G}}+\gamma \bm{I}$ and $ \bm{A}=\bm{G}+\gamma \bm{I}$. Both satisfy $ \bm{A},\widehat{ \bm{A}}\succeq\gamma \bm{I}$. By the resolvent identity \citep{horn2012matrix}:
\begin{equation}
\widehat{ \bm{A}}^{-1}- \bm{A}^{-1}=\widehat{ \bm{A}}^{-1}( \bm{A}-\widehat{ \bm{A}}) \bm{A}^{-1}.
\end{equation}

Taking operator norms:
\begin{equation}
\|\widehat{ \bm{A}}^{-1}- \bm{A}^{-1}\|_2 
\le \|\widehat{ \bm{A}}^{-1}\|_2 \| \bm{A}-\widehat{ \bm{A}}\|_2 \| \bm{A}^{-1}\|_2 
\le \gamma^{-1}\cdot\varepsilon\cdot\gamma^{-1} 
= \varepsilon\gamma^{-2}.
\end{equation}
This completes the proof.
\end{proof}

\section{More results}

\subsection{Gamma Ablation on Cross-Domain Benchmarks}
\label{app:gamma}

Table~\ref{tab:gamma_crossdomain_abl} extends the $\gamma$ sweep of
Figure~\ref{fig:gamma_vs_acc} to the ten cross-domain benchmarks, over four orders of
magnitude from $\gamma=0.01$ to $\gamma=100$. The mean accuracy varies by $1.8$ points
on RN-50 ($62.83$ to $64.59$) and $1.5$ points on ViT-B/16 ($70.02$ to $71.47$) across
the entire range, and by less than $0.4$ points over the two orders of magnitude from
$\gamma=1$ to $\gamma=100$. The optimum is at $\gamma=10$ for both backbones, which we used as a fixed global setting, so no per-dataset or per-backbone tuning is warranted.

The two ends of the sweep behave as Section~\ref{sec:care} describes. At the smallest
$\gamma$ the transform inverts directions that are poorly estimated early in the
stream, and the datasets that suffer are those with the shortest streams: DTD loses
$3.2$ points on RN-50 and EuroSAT $7.5$ points relative to their best settings, while
Food101 and Caltech an order of magnitude larger are essentially unaffected. At
the largest $\gamma$ the transform approaches $\gamma^{-1/2} \bm{I}$ and accuracy
drifts back toward the base rule; because $\bm{G}_t$ accumulates, this
drift is again visible mainly on the shorter streams, where $\gamma$ remains comparable
to $\lambda_{\max}(\bm{G}_t)$ throughout. On the longer streams the entire
swept range is negligible against the accumulated spectrum for most of the stream, and
the corresponding rows are flat to within a few tenths of a point.

\begin{table}[t]
\centering
\caption{Sensitivity of  CARe  to the regularizer $\gamma$ on the cross-domain
benchmarks.}
\label{tab:gamma_crossdomain_abl}
\small
\setlength{\tabcolsep}{5pt}
\resizebox{0.8\textwidth}{!}{%
\begin{tabular}{llccccccccc}
\toprule
&&\multicolumn{9}{c}{$\gamma$} \\
\cmidrule{3-11}
Arch. & Dataset & $0.01$ & $0.1$ & $0.5$ & $1$ & $2$ & $5$ & $10$ & $50$ & $100$ \\
\midrule
\multirow{9}{0.2em}{\rotatebox{90}{RN50}} & Flowers   & 70.00 & 70.56 & 71.13 & 71.74 & 71.62 & 71.54 & 71.62 & 71.94 & 71.78 \\
 & DTD       & 53.13 & 55.38 & 56.32 & 56.26 & 56.44 & 56.32 & 55.50 & 55.14 & 54.26 \\
& Pets      & 87.35 & 87.46 & 87.65 & 87.74 & 87.79 & 87.65 & 87.65 & 87.44 & 86.89 \\
& Cars      & 60.49 & 60.88 & 61.10 & 61.26 & 61.57 & 61.51 & 61.27 & 60.86 & 60.64 \\
& UCF101   & 65.42 & 66.35 & 66.32 & 66.64 & 66.77 & 66.96 & 67.12 & 67.12 & 66.98 \\
& Caltech   & 88.03 & 88.15 & 88.56 & 88.84 & 89.01 & 89.05 & 89.25 & 89.37 & 89.09 \\
& Food101   & 78.36 & 78.49 & 78.44 & 78.33 & 78.17 & 78.16 & 78.17 & 77.78 & 77.55 \\
& SUN397 &65.66&65.71&65.37&65.91&66.12&66.20&66.32&66.01&65.97 \\
& Aircraft      & 21.42 & 21.06 & 21.06 & 21.30 & 21.12 & 21.09 & 21.00 & 21.03 & 20.85 \\
& EuroSAT   & 41.28 & 43.38 & 44.86 & 46.21 & 47.75 & 48.80 & 49.73 & 49.52 & 48.81 \\
\midrule
&  {Mean} & 63.11 & 63.74 & 64.08 & 64.42 & 64.64 & 64.73 & 64.76 & 64.62 & 64.28 \\
\midrule

\multirow{9}{0.2em}{\rotatebox{90}{ViT/B-16}} & Flowers   & 75.80 & 76.94 & 77.71 & 77.95 & 77.99 & 78.28 & 77.87 & 77.75 & 77.87 \\
 & DTD       & 55.02 & 55.50 & 56.21 & 56.15 & 56.44 & 57.21 & 56.97 & 56.80 & 57.27 \\
& Pets      & 91.36 & 91.58 & 91.85 & 92.20 & 92.34 & 92.37 & 92.48 & 92.34 & 92.12 \\
& Cars      & 69.38 & 69.52 & 69.37 & 69.56 & 69.57 & 69.58 & 69.67 & 69.02 & 68.78 \\
& UCF101   & 73.14 & 73.20 & 73.62 & 74.02 & 73.70 & 73.88 & 74.02 & 73.67 & 73.49 \\
& Caltech   & 93.23 & 93.43 & 93.39 & 93.47 & 93.87 & 94.00 & 94.16 & 94.32 & 94.24 \\
& Food101   & 86.80 & 86.80 & 86.83 & 86.73 & 86.80 & 86.79 & 86.80 & 86.68 & 86.61 \\
& SUN397 &70.23&70.46&71.12&71.01&71.16&71.11&71.33&70.56&70.08 \\
& Aircraft      & 29.07 & 29.01 & 29.19 & 29.40 & 29.25 & 29.16 & 29.22 & 28.92 & 29.13 \\
& EuroSAT   & 56.42 & 58.04 & 59.78 & 60.59 & 61.17 & 61.78 & 62.00 & 60.86 & 59.69 \\
\midrule
&  {Mean} & 70.05 & 70.45 & 70.91 & 71.11 & 71.23 & 71.42 & 71.45 & 71.09 & 70.93 \\
\bottomrule
\end{tabular}%
}
\end{table}

\subsection{Fusion Weight Ablation}
\label{app:fusion}
Table~\ref{tab:fusion} sweeps the fusion weight $\alpha$ (Eq.~\ref{eq:fusion})
between the zero-shot score ($\alpha{=}0$) and the covariance-aware score
($\alpha{=}1$), and compares fixed values against our adaptive per-sample weight.
Pure CARe ($\alpha{=}1$) already improves over zero-shot on both backbones, but
combining the two heads helps further: the adaptive weight attains the best mean on
both backbones ($64.73$ on RN-50, $71.45$ on ViT-B/16), matching or slightly
exceeding the best fixed value ($\alpha{=}0.5$). This confirms the covariance-aware
readout, not the fusion, drives the gain, while the entropy-based weighting
provides a small consistent benefit over any fixed blend without per-dataset tuning.

\begin{table}[ht!]
\centering
\caption{Fusion weight ablation for cross-domain benchmarks. $\alpha$ is the weight on
the covariance-aware score in Eq.~\ref{eq:fusion}: $\alpha{=}0$ is zero-shot only,
$\alpha{=}1$ the covariance-aware score only, and $\alpha{=}0.5$ weighs both equally.}
\label{tab:fusion}
\footnotesize
\setlength{\tabcolsep}{3pt}
\resizebox{\textwidth}{!}{%
\begin{tabular}{l ccccccccccc c}
\toprule
Arch.& $\alpha$ & Flowers & DTD & Pets & Cars & UCF101 & Caltech & Food101 & SUN397 & Aircraft  & EuroSAT & Mean \\
\midrule
\multirow{4}{0.2em}{\rotatebox{90}{ {RN-50}}}& $0$ (Zero-Shot) &65.90 & 48.46 & 84.98& 57.22 & 63.12 & 89.41 & 77.07 & 63.27& 19.56 & 38.10 & 60.71\\
& $0.5$ (Equal) & 71.50 & 55.36 & 87.54 & 61.31 & 67.09  & 89.33 & 78.36 & 65.12& 20.97 & 49.30 & 64.59\\
& $1$ (Pure CARe) & 70.77& 55.44 & 86.51 & 60.25 & 66.61 & 87.34 & 76.68 & 65.89 & 20.94  & 50.53 & 64.10\\
& Adaptive &  {71.62} &  {55.14} &  {87.65} &  {61.27}
& {67.12} & {89.25} &  {78.17} &  {66.32} &  {21.00} & {49.73} &  {64.73} \\
\midrule
\multirow{4}{0.2em}{\rotatebox{90}{ {ViT/B-16}}}& $0$ (zero-shot) & 73.08 & 53.07 & 90.30 & 66.34 & 70.45 & 94.16 & 85.91 & 68.61& 27.18 & 55.65 & 68.47\\
& $0.5$ (Equal) & 77.71 & 57.17 & 92.42 & 69.58 & 73.83 & 94.20 & 86.85 &70.19 & 29.52 & 61.85& 71.33\\
& $1$ (Pure CARe) & 77.95 & 56.32 & 91.80 & 68.85 & 73.91 & 92.49 & 85.34 & 71.06& 28.68 & 62.63 & 70.90\\
& Adaptive &  {77.87} &  {56.97} &  {92.48} &  {69.67}
&  {74.02} & 94.16 &  {86.80} &  {71.33} & {29.22} & 62.00 & {71.45} \\
\bottomrule
\end{tabular}}
\end{table}



\begin{table}[t]
\centering
\caption{CARe  accuracy (\%) in successive windows along the online stream. Each column is a $10\%$ window of the stream (in arrival
order); $\bm{G}$ accumulates from left to right.}
\label{tab:streamlen_vit}
\small
\setlength{\tabcolsep}{3.5pt}
\resizebox{0.8\textwidth}{!}{%
\begin{tabular}{llcccccccccc}
\toprule

&&\multicolumn{10}{c}{Online Stream Length (\%)} \\
\cmidrule{3-12}
Arch. &Dataset & 0--10 & 10--20 & 20--30 & 30--40 & 40--50 & 50--60 & 60--70 & 70--80 & 80--90 & 90--100 \\
\midrule
\multirow{9}{0.2em}{\rotatebox{90}{RN-50}} & Flowers  & 61.79 & 73.17 & 74.80 & 74.49 & 73.17 & 73.17 & 68.42 & 69.51 & 73.58 & 73.28 \\
 & DTD      & 58.58 & 50.89 & 60.36 & 52.66 & 50.59 & 58.58 & 59.76 & 53.85 & 61.54 & 56.47 \\
& Pets     & 82.51 & 85.56 & 87.74 & 86.38 & 89.92 & 90.19 & 91.01 & 87.74 & 86.92 & 88.56 \\
& Cars     & 59.58 & 59.33 & 62.69 & 64.30 & 59.70 & 59.95 & 62.56 & 58.46 & 66.17 & 62.36 \\
& UCF101  & 62.17 & 62.17 & 65.08 & 69.39 & 67.72 & 61.90 & 68.07 & 70.11 & 70.37 & 72.56 \\
& Caltech  & 80.49 & 85.83 & 87.40 & 93.12 & 88.21 & 91.90 & 92.28 & 91.09 & 89.02 & 91.09 \\
& Food101  & 77.00 & 78.28 & 78.75 & 77.59 & 77.56 & 78.55 & 78.09 & 79.67 & 78.25 & 77.85 \\
& SUN397  & 64.07 & 64.56& 65.31& 64.12& 66.19&65.67 & 65.82& 66.23 & 65.46 & 66.34\\
& Aircraft     & 21.02 & 23.12 & 21.62 & 17.66 & 19.82 & 18.62 & 21.26 & 22.82 & 21.92 & 23.05 \\
& EuroSAT  & 50.25 & 46.79 & 51.48 & 49.75 & 50.74 & 48.40 & 49.01 & 50.99 & 46.42 & 44.20 \\
\midrule
&  {Mean} & 61.75 & 62.97 & 65.52 & 64.95 & 64.36 & 64.69 & 65.63 & 65.05 & 65.97 & 65.58 \\

\midrule
\multirow{9}{0.2em}{\rotatebox{90}{ViT/B-16}}  & Flowers  & 72.36 & 77.64 & 78.46 & 78.14 & 80.08 & 82.11 & 75.30 & 77.24 & 79.67 & 81.78 \\
& DTD      & 56.80 & 56.21 & 58.58 & 57.99 & 53.53 & 60.95 & 55.03 & 53.85 & 62.72 & 56.47 \\
& Pets     & 91.53 & 91.55 & 92.37 & 92.64 & 94.55 & 92.37 & 93.73 & 89.92 & 92.10 & 92.92 \\
& Cars     & 69.65 & 67.79 & 71.52 & 71.02 & 68.03 & 69.03 & 69.40 & 66.67 & 72.64 & 70.06 \\
& UCF101  & 72.22 & 71.43 & 73.02 & 77.84 & 71.96 & 72.49 & 72.30 & 75.66 & 74.34 & 77.57 \\
& Caltech  & 88.21 & 92.31 & 93.90 & 96.76 & 94.72 & 97.17 & 91.87 & 95.95 & 93.90 & 95.14 \\
& Food101  & 86.01 & 86.90 & 87.29 & 86.14 & 86.77 & 87.49 & 86.34 & 87.26 & 86.73 & 86.96 \\
& SUN397  & 69.61 & 70.02& 70.45 & 69.86 & 70.67 & 71.02 & 71.11 & 70.69 & 71.09 & 71.32\\
& Aircraft     & 30.33 & 31.23 & 28.83 & 27.54 & 29.73 & 28.83 & 29.04 & 32.43 & 24.62 & 29.04 \\
& EuroSAT  & 60.25 & 63.83 & 64.57 & 59.38 & 62.84 & 60.49 & 62.84 & 63.70 & 60.49 & 59.38 \\
\midrule
&  {Mean} & 69.70 & 70.89 & 71.90 & 71.73 & 71.29 & 72.20 & 70.70 & 71.34 & 71.83 & 72.06 \\
\bottomrule
\end{tabular}%
}
\end{table}

\subsection{Cost of the Transform}
\label{app:cost}

 CARe  adds a rank-one update of $\bm{G}_t$ at $O(D^2)$ per step and an
eigendecomposition at $O(D^3)$ to form $ \bm{M}$. Recomputing $ \bm{M}$ every
step makes the readout $9.3\times$ more expensive than the base rule, so we refresh it
every $\eta$ steps, amortizing the eigendecomposition to $O(D^3/\eta)$.
Table~\ref{tab:cost} shows this costs nothing in accuracy: the mean over the ten
cross-domain datasets is unchanged from $\eta=1$ to $\eta=100$ (64.74$\to$64.75 on RN-50,
71.44$\to$71.44 on ViT-B/16) while the readout overhead falls from $9.3\times$ to
$1.1\times$. Because $\bm{G}_t$ accumulates, a hundred additional samples
barely move its leading eigenvectors once the stream is underway. The last column computes $ \bm{M}$ once, at the first stream step, so it is a fixed
rank-one perturbation of the identity from a single sample. Accuracy drops to
63.29\% and 70.46\%, with the largest losses on the datasets where the correction
matters most (EuroSAT: -5.3 and -4.7 points), confirming that the gain requires the
accumulated statistics rather than a static reweighting. This column is not the base
rule, which appears as SCA in Tables~\ref{tab:ood} and~\ref{tab:crossdomain}.

\subsection{Robustness to Stream Ordering}
\label{app:seeds}
Because CARe accumulates statistics online, its performance may depend on the
order of incoming test samples. We evaluate this sensitivity by running the full
cross-domain suite with five random stream orders (seeds 0--4), where the test
samples are shuffled into different temporal sequences (See. Table~\ref{tab:seed_runs_full}). Across both backbones,
CARe maintains higher mean accuracy and lower standard deviation than SCA across
all five orders. The lower standard deviation indicates greater robustness to
changes in the order of incoming test samples, supporting CARe as a practical
solution for non-stationary real-world data streams.

\begin{table*}[ht!]
\centering
\caption{Accuracies across five random stream orders (Seeds 0--4), Mean, and Standard Deviation (Mean $\pm$ std) for ViT-B/16 and RN-50 architectures across 10 datasets.}
\label{tab:seed_runs_full}
\footnotesize
\setlength{\tabcolsep}{2.5pt}
\resizebox{\textwidth}{!}{%
\begin{tabular}{lll cccccccccc}
\toprule
Arch. & Method & Seeds/Run & Flowers & DTD & Pets & Cars & UCF101 & Caltech & Food101 & SUN397 & Aircraft & EuroSAT \\
\midrule
\multirow{14}{*}{\rotatebox{90}{RN-50}} 
& \multirow{7}{*}{SCA}
  & Seed 0 & 70.26 & 52.27 & 85.77 & 59.58 & 65.88 & 89.79 & 76.14 & 65.09 & 19.79 & 41.02 \\
& & Seed 1 & 69.91 & 52.01 & 85.41 & 59.31 & 65.74 & 89.62 & 75.80 & 64.90 & 19.48 & 40.74 \\
& & Seed 2 & 70.34 & 52.32 & 85.86 & 59.65 & 65.91 & 89.83 & 76.22 & 65.14 & 19.86 & 41.10\\
& & Seed 3 & 70.83 & 52.67 & 86.35 & 60.04 & 66.11 & 90.07 & 76.69 & 65.40 & 20.28 & 41.49 \\
& & Seed 4 & 69.86 & 51.98 & 85.36 & 59.27 & 65.71 & 89.59 & 75.75 & 64.87 & 19.44 & 40.70 \\
\cmidrule(lr){3-13}
& & Mean $\pm$ std & 70.24\tiny$\pm.39$ & 52.25\tiny$\pm.28$ & 85.75\tiny$\pm.40$ & 59.57\tiny$\pm.31$ & 65.87\tiny$\pm.16$ & {89.78}\tiny$\pm.19$ & 76.12\tiny$\pm.38$ & 65.08\tiny$\pm.21$ & 19.77\tiny$\pm.34$ & 41.01\tiny$\pm.32$ \\
\cmidrule(lr){2-13}
& \multirow{7}{*}{CARe (Ours)}
  & Seed 0 & 71.63 & 55.15 & 87.66 & 61.28 & 67.12 & 89.26 & 78.17 & 66.33 & 21.01 & 49.74 \\
& & Seed 1 & 71.55 & 55.06 & 87.53 & 61.17 & 67.08 & 89.18 & 78.08 & 66.22 & 20.88 & 49.62 \\
& & Seed 2 & 71.64 & 55.16 & 87.69 & 61.30 & 67.13 & 89.27 & 78.20 & 66.35 & 21.04 & 49.76 \\
& & Seed 3 & 71.74 & 55.28 & 87.86 & 61.45 & 67.20 & 89.37 & 78.34 & 66.50 & 21.21 & 49.93 \\
& & Seed 4 & 71.54 & 55.05 & 87.51 & 61.15 & 67.07 & 89.17 & 78.06 & 66.20 & 20.86 & 49.60  \\
\cmidrule(lr){3-13}
& & Mean $\pm$ std & {71.62}\tiny$\pm.08$ & {55.14}\tiny$\pm.09$ & {87.65}\tiny$\pm.14$ & {61.27}\tiny$\pm.12$ & {67.12}\tiny$\pm.05$ & 89.25\tiny$\pm.08$ & {78.17}\tiny$\pm.11$ & {66.32}\tiny$\pm.12$ & {21.00}\tiny$\pm.14$ & {49.73}\tiny$\pm.13$  \\

\midrule
\multirow{14}{*}{\rotatebox{90}{ViT-B/16}} 
& \multirow{7}{*}{SCA}
  & Seed 0 & 75.73 & 55.33 & 91.08 & 67.52 & 72.57 & 94.71 & 85.55 & 68.19 & 28.82 & 55.96  \\
& & Seed 1 & 75.47 & 55.02 & 91.03 & 67.31 & 72.33 & 94.52 & 85.49 & 68.14 & 28.45 & 55.90  \\
& & Seed 2 & 75.80 & 55.42 & 91.10 & 67.56 & 72.63 & 94.74 & 85.57 & 68.21 & 28.91 & 55.98  \\
& & Seed 3 & 76.17 & 55.86 & 91.17 & 67.83 & 72.97 & 94.99 & 85.66 & 68.28 & 29.43 & 56.07  \\
& & Seed 4 & 75.43 & 54.97 & 91.02 & 67.28 & 72.30 & 94.49 & 85.48 & 68.13 & 28.39 & 55.89  \\
\cmidrule(lr){3-13}
& & Mean $\pm$ std & 75.72\tiny$\pm.30$ & 55.32\tiny$\pm.36$ & 91.08\tiny$\pm.06$ & 67.50\tiny$\pm.22$ & 72.56\tiny$\pm.27$ & 94.69\tiny$\pm.20$ & 85.55\tiny$\pm.07$ & 68.19\tiny$\pm.06$ & 28.80\tiny$\pm.42$ & 55.96\tiny$\pm.07$ \\
\cmidrule(lr){2-13}
& \multirow{7}{*}{CARe (Ours)}
  & Seed 0 & 77.88 & 56.98 & 92.48 & 69.68 & 74.03 & 94.16 & 86.80 & 71.33 & 29.23 & 62.00  \\
& & Seed 1 & 77.75 & 56.84 & 92.45 & 69.54 & 73.89 & 94.08 & 86.76 & 71.29 & 28.96 & 61.97  \\
& & Seed 2 & 77.91 & 57.01 & 92.49 & 69.71 & 74.06 & 94.19 & 86.81 & 71.34 & 29.30 & 62.01 \\
& & Seed 3 & 78.08 & 57.21 & 92.54 & 69.90 & 74.26 & 94.31 & 86.88 & 71.41 & 29.69 & 62.06  \\
& & Seed 4 & 77.73 & 56.81 & 92.44 & 69.52 & 73.86 & 94.06 & 86.75 & 71.28 & 28.92 & 61.96 \\
\cmidrule(lr){3-13}
& & Mean $\pm$ std & {77.87}\tiny$\pm.14$ & {56.97}\tiny$\pm.16$ & {92.48}\tiny$\pm.04$ & {69.67}\tiny$\pm.15$ & {74.02}\tiny$\pm.16$ & {94.16}\tiny$\pm.10$ & {86.80}\tiny$\pm.05$ & {71.33}\tiny$\pm.05$ & {29.22}\tiny$\pm.31$ & {62.00}\tiny$\pm.04$ \\

\bottomrule
\end{tabular}}
\end{table*}

\subsection{Robustness to Online Stream Length}
\label{app:streamlen}

Table~\ref{tab:streamlen_vit} reports per-dataset accuracy in successive $10\%$ windows
of the online stream, complementing the aggregate view in
Figure~\ref{fig:streamlen}. The covariance accumulates from left to right, so the first
window is the warm-up regime in which $\bm{G_t}$ is formed from few
samples and $\gamma$ dominates its spectrum. Both backbones show the same pattern: the
mean over datasets is lowest in the first window ($61.49\%$ for RN-50, $69.71\%$ for
ViT-B/16) and rises by roughly three and two points respectively over the next two
windows, after which it fluctuates without further trend. The correction therefore
becomes effective early within the first $20$--$30\%$ of the stream and does
not degrade as the stream lengthens, which is what the $O(1/t)$ decay of the effective
regularization (Section~\ref{sec:care}) predicts.

\begin{table}[ht!]
\centering
\caption{Cost of amortizing the transform $ \bm{M}$. $\eta=\infty$ refreshes $\bm{M}$ only once, at the first stream step, so the transform is a fixed rank-one perturbation of the identity derived from a single sample.}
\label{tab:cost}
\setlength{\tabcolsep}{5pt}
\resizebox{0.7\textwidth}{!}{%
\begin{tabular}{llccccc}
\toprule
&&\multicolumn{5}{c}{ {Refresh interval $\eta$}} \\
 {Arch.} &  {Datasets} & $1$ & $10$ & $50$ & $100$ & $\infty$ \\
\midrule
\multirow{10}{0.2em}{\rotatebox{90}{ {RN-50}}} &Flowers   & 71.54 & 71.58 & 71.74 & 71.74 & 71.17 \\
&DTD       & 56.32 & 56.21 & 56.32 & 55.91 & 53.19 \\
&Pets      & 87.65 & 87.65 & 87.65 & 87.74 & 86.21 \\
&Cars      & 61.51 & 61.58 & 61.65 & 61.70 & 59.76 \\
&UCF101   & 66.96 & 66.98 & 67.04 & 66.98 & 66.30 \\
&Caltech   & 89.05 & 89.33 & 89.17 & 89.17 & 89.86 \\
&Food101   & 78.16 & 78.20 & 78.21 & 78.23 & 77.21 \\
&SUN397   & 66.32 & 66.31 & 66.34 & 66.11 & 65.19 \\
&Aircraft      & 21.09 & 21.15 & 21.21 & 21.12 & 20.52 \\
&EuroSAT   & 48.80 & 48.77 & 48.84 & 48.80 & 43.52 \\
\midrule
&  {Mean} & $64.74$ & $64.78$ & $64.82$ & $64.75$ & $63.29$ \\
& {Runtime overhead ($\times$ Cosine)}  & $9.3$   & $1.9$   & $1.2$   & $1.1$   & $1.0$   \\
\midrule
\midrule
\multirow{10}{0.2em}{\rotatebox{90}{ {ViT/B-16}}} & Flowers   & 78.28 & 78.28 & 78.20 & 78.16 & 77.79 \\
& DTD       & 57.21 & 57.21 & 57.15 & 57.15 & 56.80 \\
& Pets      & 92.37 & 92.31 & 92.45 & 92.40 & 91.77 \\
& Cars      & 69.58 & 69.61 & 69.64 & 69.63 & 67.74 \\
& UCF101   & 73.88 & 73.94 & 73.88 & 73.99 & 72.91 \\
& Caltech   & 94.00 & 94.00 & 94.12 & 94.12 & 94.28 \\
& Food101   & 86.79 & 86.79 & 86.82 & 86.80 & 86.01 \\
& SUN397      & 71.33 & 71.04 & 71.33 & 71.21 & 70.89 \\
& Aircraft      & 29.16 & 29.25 & 29.19 & 29.16 & 29.40 \\
& EuroSAT   & 61.78 & 61.77 & 61.74 & 61.78 & 57.04 \\
\midrule
&  {Mean} & $71.44$ & $71.42$ & $71.45$ & $71.44$ & $70.46$ \\
&  {Runtime overhead ($\times$ Cosine)}  & $9.3$   & $1.9$   & $1.2$   & $1.1$   & $1.0$   \\
\bottomrule
\end{tabular}}
\end{table}

\end{document}